\documentclass[accepted]{uai2024} 
                        
\usepackage[american]{babel}

\usepackage{natbib} 
\usepackage{mathtools} 
\usepackage{booktabs} 
\usepackage{tikz} 
\usepackage{amssymb}
\usepackage{amsthm}
\usepackage{algorithm}
\usepackage{algorithmic}

\usepackage{multirow}

\newtheorem{theorem}{Theorem}[section]
\newtheorem{proposition}[theorem]{Proposition}

\newtheorem{definition}[theorem]{Definition}
\newtheorem{assumption}[theorem]{Assumption}
\newtheorem{problem}[theorem]{Problem}

\title{Disentangled and Distilled Encoder for Out-of-Distribution Reasoning with Rademacher Guarantees}

\author[1]{\href{mailto:<jj@example.edu>?Subject=Your UAI 2024 paper}{Zahra Rahiminasab}{}}
\author[1]{ Michael Yuhas}
\author[1,2]{Arvind Easwaran
 }

\affil[1]{%
    School of Computer Science
and Engineering\\
Nanyang Technological University\\
Singapore, Singapore\\
}
\affil[2]{%
    Energy Research Institute\\
Nanyang Technological University\\
Singapore, Singapore\\
}
\affil[3]{%
    School of Computer Science
and Engineering\\
Nanyang Technological University\\
Singapore, Singapore\\
}
  
\begin{document}
\maketitle

\begin{abstract}
Recently, the disentangled latent space of a variational autoencoder (VAE) has been used to reason about multi-label out-of-distribution (OOD)  test samples that are derived from different distributions than training samples. Disentangled latent space means having one-to-many maps between latent dimensions and generative factors or important characteristics of an image. This paper proposes a disentangled distilled encoder (DDE) framework to decrease the OOD reasoner size for deployment on resource-constrained devices while preserving disentanglement. DDE formalizes student-teacher distillation for model compression as a constrained optimization problem while preserving disentanglement with disentanglement constraints. Theoretical guarantees for disentanglement during distillation based on Rademacher complexity are established. The approach is evaluated empirically by deploying the compressed model on an NVIDIA Jetson Nano.
\end{abstract}

\section{Introduction}
Deep learning  (DL)  models may make incorrect predictions with high confidence when they receive out-of-distribution (OOD) test samples that are derived from different distributions than training samples. Presence of OOD samples is dangerous in safety-critical cyber-physical systems (CPS) such as autonomous vehicles (AV), where wrong predictions for these samples can lead to fatal results. To address this issue, the decision manager unit is designed to receive outputs of the DL model and OOD detector at inference time to determine the reliability of the DL model's predicted results based on the OOD detector outcome. OOD reasoning focuses on identifying the source of OOD behavior based on generative factors. Generative factors like brightness are important for describing an image ~\citep{plumerault2019controlling}. 
Identifying the source of OOD behavior helps to identify proper safe-fail mechanisms, such as returning control to a human driver.

A variational autoencoder (VAE)  architecture includes an encoder, a decoder, and a latent space. The encoder maps data to a lower-dimensional latent space before the decoder reconstructs the input by sampling latent space. Data distribution is learned in the latent space ~\citep{goodfellow2016deep} by simultaneously training the encoder and decoder. Although OOD analysis in a VAE's output space is error-prone ~\citep{nalisnick2019detecting}, using its latent space for detecting OOD samples shows promising results for single label ~\citep{vasilev2020q,zhang2020towards} and multi-label data  ~\citep{ramakrishna2022efficient}. A  latent space of VAE must be disentangled for interpretable OOD reasoning results, where each latent dimension mostly represents one generative factor.

Resource-constrained safety-critical CPS like Jetson Nano ~\citep{9126102} share resources like CPU between the DL model and OOD reasoner. Thus, the OOD reasoner model must be small and have a short inference time to meet hard deadlines in such CPS ~\citep{cai2020real}. Although knowledge distillation ~\citep{gou2021knowledge} can compress a deeper OOD reasoner to a shallower one with fewer neurons, it is important to preserve disentanglement during distillation to maintain OOD reasoner performance.

Current solutions for disentanglement during distillation use constrained optimization and focus on the domain generalization (DG) problem. 
In DG,  knowledge distillation is used to disentangle objects and background by separating them with different models ~\citep{robey2021model,zhang2022towards} rather than using knowledge distillation to compress a given model while preserving disentanglement. So, these approaches are suitable for single-label data and require multiple disentanglement models, making them resource-intensive and infeasible for CPS. Also, they are fully supervised, or a subset of samples are supervised (restricted labeling) and make assumptions about the ideal teacher model ~\citep{cha2022domain}.

This paper presents a disentangled distilled encoder (DDE) for multi-label data that compresses the OOD reasoner while preserving disentanglement. Training is formulated as a constrained optimization problem by adapting the approach in  ~\citep{chamon2022constrained}. DDE uses knowledge distillation to compress the teacher model with more neurons to the student model with fewer neurons. Disentanglement is preserved by enforcing   \textit{Adaptability} and \textit{Isolation} constraints. \textit{Adaptability} means information about a change in a generative factor in representative dimensions is transferred from the teacher to the student model. \textit{Isolation} means the gap between the average mutual information defined over representative and unrepresentative latent dimensions for a given factor is preserved during knowledge distillation from the teacher to the student. In contrast to previous approaches, DDE is weakly supervised with match-pairing, i.e., only groups of samples with the same value for a given factor are available during training ~\citep{shu2019weakly}. It can be used with any model that can partially disentangle the multi-label data, i.e., total disengagement is impossible in practice due to unknown generative factors. It does not require any information regarding OOD samples during training. 

We analyze the optimality of solutions for a constraint optimization problem based on parameterization and empirical gaps~\citep{chamon2022constrained}. The parameterization gap occurs when a non-convex deep model (like an encoder) is used to learn a convex learning task (like feature extraction). An empirical gap arises because deep learning models only have access to training samples during training rather than the entire input space. We evaluate both gaps and utilize the Rademacher complexity (RC) ~\citep{mohri2018foundations} of the model to limit the expected loss functions. In summary, we make the following contributions:
\begin{itemize}
    \item  We formalized the training of a weakly disentangled distilled student model with a smaller size than the teacher model as a constraint optimization problem.
    \item We analyze the optimality of the obtained solutions for a defined problem based on parameterization and empirical gaps by adapting the theoretical results in ~\citep{chamon2022constrained} to match-pairing supervision. We bound the expectation of defined loss functions based on  RC.
    \item We empirically show the preservation of OOD performance by a student model trained on the CARLA dataset~\citep{Dosovitskiy2017} and evaluated on a Jetson Nano.
\end{itemize}


\section{Related work}
\label{RelatedWork}
Knowledge distillation is commonly used to achieve disentanglement in domain generalization (DG) and information bottleneck (IB) problems. DG focuses on training a model in one domain while obtaining acceptable performance in an unseen domain at run-time~\citep{zhou2021domain}. IB aims to learn a compressed data representation that maintains important data characteristics~\citep{pan2021disentangled}. Disentanglement refers to the complete separation of domain-specific (style) and domain-independent (content) features in these problems. Therefore, they are unsuitable for multi-label image data in which some generative factors cannot be separated completely.   Previous studies have distilled content and style into separate encoders using mutual information between data labels and content~\citep{pan2021disentangled, yang2022factorizing} or image reconstruction based on labels~\citep{xiang2021disunknown}. These approaches require labeled data during training and, except for~\citep{wang2022disentangled}, use separate encoders for style and content features, making them resource-intensive. Finally, they do not use knowledge distillation to compress a model.

In ~\citep{robey2021model,zhang2022towards,cha2022domain}, DG  is defined as a min-max optimization problem with disentanglement constraints. Examples of such constraints are the insensitiveness of the trained network to changes in style factor~\citep{robey2021model},  consistency of the reconstructed image with changing style factor but fixed content~\citep{zhang2022towards},  or equality of ideal and domain model losses~\citep{cha2022domain}. Of these approaches, only ~\citep{robey2021model} and ~\citep{zhang2022towards} analyze the optimality of the defined problems. All three approaches solve domain generalization problems for single-label data, and all are supervised. 


\section{Disentangled Distilled Encoder (DDE)}

Our framework aims to distill a smaller student encoder $\mathcal{E}_{s} \in \mathcal{H}_s: \Theta_s \times\mathcal{X} \longrightarrow {Z}_{s}$  with $\Theta_s$  parameter space  and $Z_{s}$ latent space  from a pre-trained teacher encoder $\mathcal{E}_{\tau} \in \mathcal{H}_{\tau}: \Theta_{\tau} \times\mathcal{X} \longrightarrow {Z}_{\tau}$  with  $\Theta_{\tau}$ parameter space and ${Z}_{\tau}$ latent space while preserving disentanglement. Here, $\mathcal{X} $ is input space and $\mathcal{H}_{s}$ and $\mathcal{H}_{\tau}$ are student and teacher hypothesis spaces, respectively. Figure 1 illustrates the three phases of our framework: data partitioning, training OOD reasoners as constrained optimization, and run-time OOD reasoning. In the following subsections, each step is explained in detail.

\begin{figure}[t!]
\centering
\includegraphics[width=0.9\columnwidth]{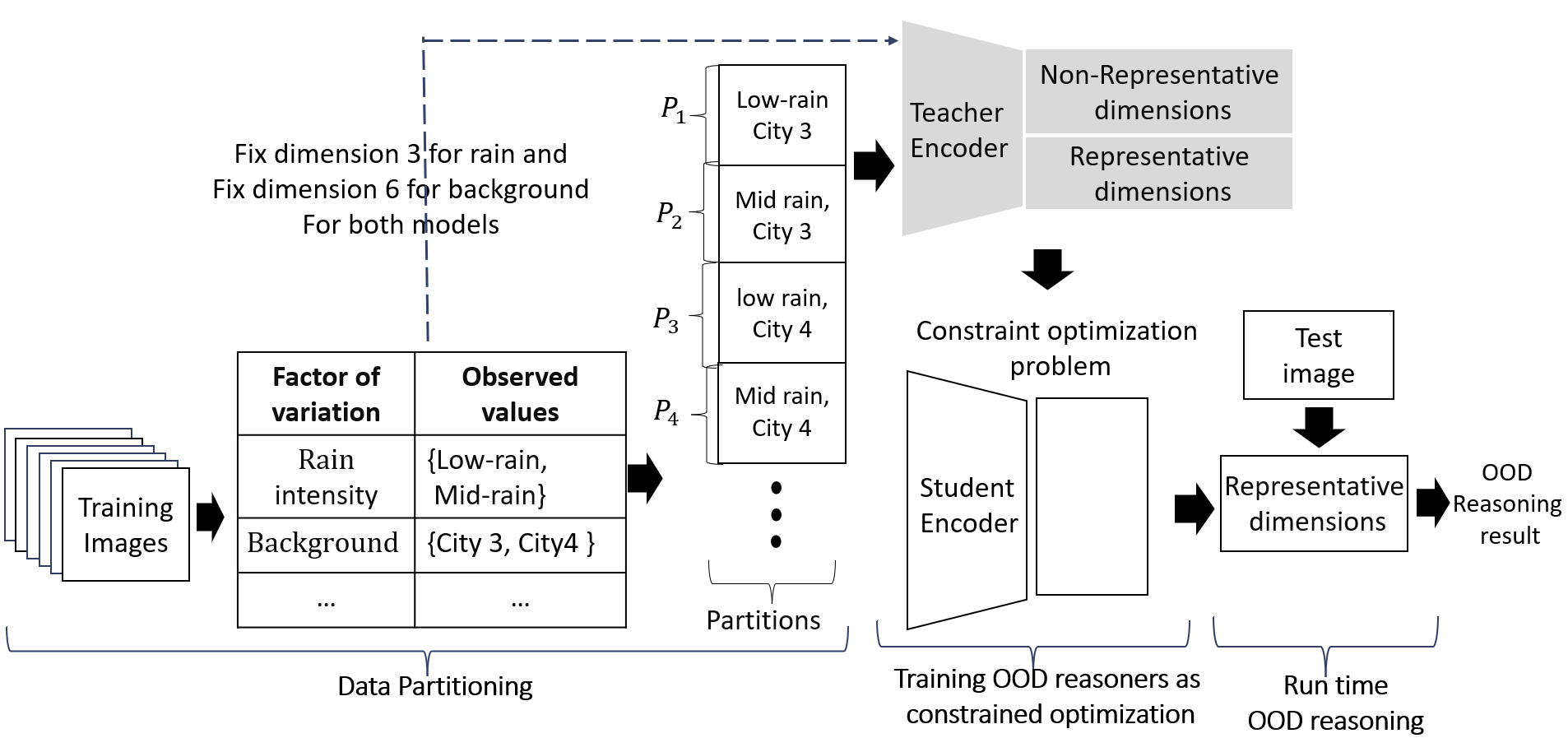}
\caption{Overview of the DDE.}
\label{fig:Distilloverall}
\end{figure}

\subsection{Data Partitioning}
\label{Datapart}
We partition the training samples $x \in \mathcal{X}_T$ based on generative factors. Each partition $P=\{x \in \mathcal{X}_T \vert (x_{1},...,x_{G})=(o^1_{b},...,o^G_{e}) \}$, where $o^{i}_j$ is the $j^{th}$  observed value for generative factor  $f_i \in \mathcal{F}=\{f_1,...,f_G\}$. The total number of partitions is defined as a combination of observed values for observed generative factors ( $K =|O_{f_1}| \times ... \times |O_{f_G}|$). $\mathcal{V}_{i}= \{(P, P') \in \mathcal{P} \times \mathcal{P}\}$  includes all pairs of partitions where the value of one generative factor $f_i$ changes, while changes in other factors are insignificant.

\subsection{Training OOD reasoners as constrained optimization problem}\label{sec: TR}

This section presents three steps to form OOD reasoners training as a constraint optimization problem:  presenting the assumptions and characteristics of the teacher model, designing student architecture, and defining main and constraint losses to form a constraint optimization problem.

The teacher model must be disentangled or partially disentangled as stated in Assumption ~\ref{ass:TechDist} to preserve disentanglement.
\begin{assumption}[Disentangled teacher model]
There is a teacher encoder $\mathcal{E}_{\tau}$ such that for given generative factor $f \in \mathcal{F}$ specific dimensions of its latent space  $\{z_j \in Z^{\tau} | j \in \mathcal{Z}^{\tau}_{f}\}$  are sensitive to changes in factor $f$, and any change in factor $f$  is isolated in these latent dimensions.   
\label{ass:TechDist}
\end{assumption}
Each learning task can be specified by a convex function in a function space. For example, $C_{\tau}$ is a convex disentangled feature extractor. Then, the non-convex hypothesis, such as the teacher encoder model $\mathcal{E}_{\tau}$, tries to cover the output of this convex function. The degree of complexity of the model identifies its ability to cover the output of corresponding convex function as shown in Assumption ~\ref{as:Compteach}.
 \begin{assumption}[Complexity of teacher hypothesis space]
Consider a closed convex hull  $\overline{\mathcal{H}_{\tau}}$ that contains all the convex hypotheses from the teacher  hypothesis space $\mathcal{H}_{\tau}$.  Then there exists $\epsilon_{\tau} \geq 0$ and $\theta_{\tau} \in \Theta_{\tau}$:
\begin{equation}
        \forall C_{\tau} \in \overline{\mathcal{H}_{\tau}}:  \;  E_{\mathfrak{D}(x)} [\vert \ C_{\tau}(x)-\mathcal{E}_{\tau}(\theta_{\tau},x) \vert] \leq \epsilon_{\tau}
    \end{equation}
     Here  $\mathfrak{D} (x)$ is the data distribution. 
     \label{as:Compteach}
\end{assumption}

For designing a student model with smaller model size, a predefined ratio of neurons from the convolution and linear layers of the teacher model are removed.   Batch normalization layers must be eliminated to avoid memory overhead. However, due to the importance of batch normalization layers in smoothing loss functions ~\citep{brock2021high},  normalization and convolution operations are combined in a convolution layer based on the approach suggested by~\citep{brock2021characterizing}. Consider a weight matrix  $W_{\beta,\alpha}$ for a layer with  $\alpha$ inputs and $\beta$ outputs. Normalized weight is defined as follows.
\begin{equation}
    \hat{W}_{\beta,\alpha} = \Gamma * \frac{W_{\beta,\alpha}-\mu_{W}}{\sigma_{W}*\sqrt{\alpha}}
\end{equation}
Here, $\Gamma$ is the gain coefficient that normalizes the variance of the layer weights to be close to one. Also,    $\mu_{W}= \frac{1}{\alpha} \sum_{j=1}^{\alpha} W_{\beta,j} $ and $\sigma_{W}=\frac{1}{\alpha} \sum_{j=1}^{\alpha} W_{\beta,j}^2-\mu_{W}^2$ are the average and variance over input dimensions, respectively. 

Next, we need to define the main objective and constraints. The main objective of disentanglement distillation is to ensure that the distribution of latent space is preserved during distillation.
\begin{definition}[Distillation loss]
Distillation loss  $\mathcal{L}^{\circ}_{D}$ measures the similarity between the latent space distributions learned  by the teacher and the student encoders:
\begin{equation}
\begin{aligned}
      &\mathcal{L}^{\circ}_{D}\triangleq \textsc{JS}(\mathcal{E}_{\tau}(\theta_{\tau},x),\mathcal{E}_s(\theta_s,x))= \\
    &\frac{1}{2}(\textsc{KL}(\mathcal{E}_{\tau}(\theta_{\tau},x),\mathcal{E}_s(\theta_s,x))+\textsc{KL} (\mathcal{E}_s(\theta_s,x), \mathcal{E}_{\tau} (\theta_{\tau},x)) =\\
    & -\frac{1}{2N}\sum_{k=1}^{N}[((ln\sigma^{\tau}_{k}- ln\sigma^{s}_{k}) - \frac{(e^{ln\sigma^{\tau}_{k}} + (\mu^{\tau}_{k}- \mu^{s}_{k}))^2 }{ e^{ln\sigma^{s}_{k}}}) \\ & +((ln\sigma^{s}_{k} - ln\sigma^{\tau}_{k}) - \frac{(e^{ln\sigma^{s}_{k}} + (\mu^{s}_{k}- \mu^{\tau}_{k}))^2 }{ e^{ln\sigma^{\tau}_{k}}} ) +2]
   \end{aligned}
    \label{eq:disloss}
       \end{equation}
Here, \textsc{JS} and \textsc{KL} are Jensen-Shannon and Kullback–Leibler divergences. \textsc{JS}  is a symmetric distance metric between two distributions, bounded by   $1$ ~\citep{lin1991divergence}. Also,  $ln\sigma^{\tau}$ and   $\mu^{\tau}$, $ln\sigma^{s}$ and  $\mu^{s}$ are the logarithm of variances and means of distributions learned by the teacher and the student encoders. In addition $|{Z}^{\tau}_{f}|=|\mathcal{Z}^{s}_{f}|=N$  is the size of latent space.
\label{df:convm}
\end{definition}
By defining and enforcing disentanglement constraints, disentanglement is preserved during distillation. Both disentanglement constraints are defined based on information change between input samples and student latent representations. However, the information function is not differentiable with respect to student model parameters. Therefore, a differentiable form of mutual information ~\citep{cha2022domain} is used.   Also, as the input sample cannot be used directly,  instead of input, the teacher representation is used to measure information change following a similar approach to ~\citep{cha2022domain}. Thus, the probability of observing a sample generated from the teacher model by the distribution of the student model is evaluated.
Given teacher distribution $\mathcal{N}(\mu^{\tau}, ln \sigma^{\tau})$ and student distribution $\mathcal{N}(\mu^{s}, ln \sigma^{s})$  for input $x$ in a mini-batch,  sample $a=\varepsilon *\sigma^{\tau}+\mu^{\tau}$  is derived from the teacher latent distribution in a given dimension. Mutual information is defined as follows:
\begin{equation}
\begin{aligned}
     I(a,\mu^s, ln\sigma^s )= \frac{-1}{2}[ln\sigma^s+\frac{(a-\mu^s)^2}{e^{ln\sigma^s}}],
\end{aligned}  
\label{eq:muins}
\end{equation}

\begin{definition}[Disentanglement for student model]
     Consider a pair of training samples  $(x,x') \in (P, P') \in \mathcal{V}_f$ that differ only in value for generative factor  $f$ and mutual information function  $I$ that is defined in Equation ~\ref{eq:muins}.
Suppose Assumption ~\ref{ass:TechDist} holds for the teacher and student latent dimensions where indexes  $\mathcal{Z}^{s}_{f}$  represent factor $f$.  We can define  adaptability and isolation constraints to preserve disentanglement by considering the same latent space size and representative dimensions for teacher and student models as follows:
\begin{itemize}[wide=0pt]
        \item \textbf{Adaptability:} 
        The adaptability constraint ensures that whenever factor  $f$ changes, the teacher's information about the changed factor is transferred to the student model in representative dimensions with indexes $ \mathcal{Z}^s_{f}$.
        \begin{equation}
        \begin{aligned}
             &\forall (x,x') \in \mathcal{V}_{f}, \forall k \in \mathcal{Z}^{s}_{f}: \\
             &I(\textsc{SM}^{\tau}_{k}(\mathcal{E}_{\tau}(\theta_{\tau},x)),\textsc{MN}^s_k(\mathcal{E}_{s}(\theta_{s},x)),\textsc{LV}^s_k(\mathcal{E}_{s}(\theta_{s},x)))= \\ &
             I(\textsc{SM}^{\tau}_{k}(\mathcal{E}_{\tau}(\theta_{\tau},x')),\textsc{MN}^s_k(\mathcal{E}_{s}(\theta_{s},x')),\textsc{LV}^s_k(\mathcal{E}_{s}(\theta_{s},x')))
        \end{aligned} 
        \label{eq:adapt}
        \end{equation}
        Here, $\textsc{SM}^{\tau}$ is a function that returns samples that are derived from the teacher latent distribution. Also, $\textsc{MN}^{\tau}$ and $\textsc{LV}^{\tau}$ are functions that return  mean and the logarithm of variance from outputs of student encoder.
        The differentiable form of the above constraint is defined in Equation~\ref{eq:5}. 
        \begin{equation}
            \begin{aligned}
                &\mathcal{L}^{\circ}_{A,f} (x,x')\triangleq \frac{-1}{2}*\frac{1}{|\mathcal{Z}^{s}_{f}|}\sum_{k \in \mathcal{Z}^s{f}} [(ln\sigma^{s}_{k}+\frac{(a^{\tau}_{k}-\mu^{s}_{k})^2}{e^{ln\sigma^{s}_{k}}})  \\
                &- (ln\sigma'^{s}_{k}+\frac{(a'^{\tau}_{k}-\mu'^{s}_{k})^2}{e^{ln\sigma'^{s}_{k}}})]
            \end{aligned}
            \label{eq:5}
        \end{equation}
        Here,  $a^{\tau}_{k}$ and   $a'^{\tau}_{k}$ are   $k^{th}$ dimensions of the  outputs of function \textsc{SM} for  $x$ and $x'$, respectively. Also,  $ln\sigma^{s}_{k}$,   $ln\sigma'^{s}_{k}$ are $k^{th}$ dimensions of  outputs of  \textsc{LV},  and  $\mu^{s}_{k}$, $\mu'^{s}_{k}$ are $k^{th}$ dimensions of  outputs of \textsc{MN} for  $x$ and $x'$, respectively.
        \item\textbf{Isolation:} The isolation constraint ensures that a change in factor  $f$ does not lead to an information change in non-representative dimensions with indexes  $\overline{ \mathcal{Z}^s_{f}}$. In other words, the information gap is preserved between representative and non-representative dimensions during distillation.
        \begin{equation}
         \begin{aligned}
            &\forall (x,x') \in \mathcal{V}_{f},  \forall k \in \mathcal{Z}^{s}_{f}, \forall t \in \overline{\mathcal{Z}^s_{f}}: \\
            &
            I(\textsc{SM}^{\tau}_{k}(\mathcal{E}_{\tau}(\theta_{\tau},x)),\textsc{MN}^s_k(\mathcal{E}_{s}(\theta_{s},x)),\textsc{LV}^s_k(\mathcal{E}_{s}(\theta_{s},x)))-\\&
             I(\textsc{SM}^{\tau}_{k}(\mathcal{E}_{\tau}(\theta_{\tau},x')),\textsc{MN}^s_k(\mathcal{E}_{s}(\theta_{s},x')),\textsc{LV}^s_k(\mathcal{E}_{s}(\theta_{s},x')))= \\
             &I(\textsc{SM}^{\tau}_{t}(\mathcal{E}_{\tau}(\theta_{\tau},x)),\textsc{MN}^s_{t}(\mathcal{E}_{s}(\theta_{s},x)),\textsc{LV}^s_{t}(\mathcal{E}_{s}(\theta_{s},x)))-
            \\&
            I(\textsc{SM}^{\tau}_{t}(\mathcal{E}_{\tau}(\theta_{\tau},x')),\textsc{MN}^s_{t}(\mathcal{E}_{s}(\theta_{s},x')),\textsc{LV}^s_{t}(\mathcal{E}_{s}(\theta_{s},x')))
            \end{aligned}
            \label{eq:Iso}
        \end{equation}
The differentiable form of the above constraint is defined as loss function  $\mathcal{L}^{\circ}_{I,f}$ in Equation~\ref{eq:7}.
  \begin{equation}
    \begin{aligned}
        & \mathcal{L}^{\circ}_{I,f}(x,x') \triangleq\\&   -\frac{1}{2} [\frac{1}{|\mathcal{Z}^{s}_{f}|}\sum_{k \in \mathcal{Z}^{s}_{f}}[(ln\sigma^{s}_{k}+\frac{(a^{\tau}_{k}-\mu^{s}_{k})^2}{e^{ln\sigma^{s}_{k}}}) - \\ & (ln\sigma'^{s}_{k}+\frac{(a'^{\tau}_{k}-\mu'^{s}_{k})^2}{e^{ln\sigma'^{s}_{k}}})]  +\\& \frac{1}{N-|\mathcal{Z}^{s}_{f}|}  \sum_{t \in \overline{ \mathcal{Z}^{s}_{f}}}[({ln\sigma'^{s}_{t}} + \frac{(a'^{\tau}_{t}-{\mu'^{s}_{t}})^2}{e^{{ln\sigma'^{s}_{t}}}}) - \\& ({ln\sigma^{s}_{t}}+ \frac{(a^{\tau}_{t}-{\mu^{s}_{t}})^2}{e^{{ln\sigma^{s}_{t}}}})] ] 
    \end{aligned}
    \label{eq:7}
    \end{equation}
    \end{itemize}
    \label{df:stdisent}
\end{definition}
Loss functions must be Lipschitz continuous with a bounded range to provide theoretical guarantees. 
\begin{assumption}[Lipschitz and bounded  loss functions]\label{as:LIPBOUND}
Although defined losses are not  Lipschitz and bounded in the domain of all real numbers, they can be Lipschitz continuous with a bounded range in a bounded domain.
 Therefore, we composite  \textsc{JS} from $\mathcal{L}^{\circ}_D$ with the (\textsc{SF}) ~\citep{bridle1990probabilistic}  and \textsc{I} from  $\mathcal{L}^{\circ}_A$ and $\mathcal{L}^{\circ}_I$ with the inverse of the tangent function (\textsc{AT}) ~\citep{wild1947arctangent}  and obtain  $\mathcal{L}^{\diamond}_D, \mathcal{L}^{\diamond}_A$  and  $\mathcal{L}^{\diamond}_I$ losses, respectively. 
\end{assumption}
Based on the main objective and disentanglement constraints, training of disentangled distilled encoder is formalized as follows.
\begin{problem} [Disentanglement distillation constrained optimization (DDCO)]
Consider a set of data partitions  $\mathcal{P}=\{P_1,...,P_K\}$, where
 $m=|P_1|+ ...+ |P_K|$ and $m_A=m_I=|\mathcal{V}_f|$.
Define:$\mathcal{L}_{D}^{\bullet}(\theta)\triangleq  \frac{1}{m} \sum_{i=1}^{m} \mathcal{L}^{\diamond}_{D}(x_i)$, 
        $\mathcal{L}_{A,f}^{\bullet}(\theta) \triangleq  \frac{1}{m_A} \sum_{i=1}^{m_A}  \mathcal{L}^{\diamond}_{A,f} (x_i,x'_i) $ and
        $\mathcal{L}_{I,f}^{\bullet}(\theta) \triangleq \frac{1}{m_I} \sum_{i=1}^{m_I} \mathcal{L}^{\diamond}_{I,f} (x_i,x'_i)$.    
  Then, the disentanglement distillation constrained optimization problem is defined as follows: 
\begin{equation}
    \begin{aligned}
        &min_{\theta_s \in \Theta_s}& \mathcal{L}_{D}^{\bullet}(\theta_s)\\
        &subject \; to:  &\mathcal{L}_{A,f}^{\bullet}(\theta_s) =0 \;(\forall f \in \mathcal{F})\\
        &&\mathcal{L}_{I,f}^{\bullet}(\theta_s) =0 \; (\forall f \in \mathcal{F})
    \end{aligned}
\end{equation}
\label{P:OrgDDS}
\end{problem}
Achieving complete constraint satisfaction while minimizing the main objective is impossible as the non-convex encoder model tries to cover the convex function that describes the learning task. So, a relaxed version of Problem~\ref{P:OrgDDS} with marginal satisfaction of constraints is presented as follows.
\begin{problem}[Relaxed DDCO]
Consider  $\gamma_{A,f}$  and  $\gamma_{I,f}$ as margins for the satisfaction of adaptation and isolation constraints for generative factor  $f$. Then, the relaxed DDCO problem is defined as follows:
\begin{equation}
    \begin{aligned}
        &p^* \triangleq & min_{\theta_s \in \Theta_s}  \mathcal{L}_{D}^{\bullet}(\theta_s) \\
        &subject \; to: &  \\
        &&\mathcal{L}_{A,f}^{\bullet}(\theta_s) \leq \gamma_{A,f} \;(\forall f \in \mathcal{F})\\
        &&\mathcal{L}_{I,f}^{\bullet}(\theta_s) \leq \gamma_{I,f} \; (\forall f \in \mathcal{F})
    \end{aligned}
\end{equation}
Here,  $p^*$ is an optimal solution for this problem. Since solving a constrained problem is a non-trivial task, we define a dual unconstrained problem based on the Lagrangian ~\citep{boyd2004convex} as follows.
\begin{equation}
    \begin{aligned}
       &  d^* \triangleq max_{\{\lambda_{A,f}, \lambda_{I,f}\}_{f \in \mathcal{F}}} min_{\theta_s \in \Theta_s}  \mathcal{L}_{D}+ \\ & \sum_{f \in \mathcal{F}}
         [\lambda_{A,f} *\mathcal{L}_{A,f} +
         \lambda_{I,f} *\mathcal{L}_{I,f} ]
    \end{aligned}
\end{equation}
Here,  $d^*$ is an optimal solution;  $\mathcal{L}_{D}=\mathcal{L}_{D}^{\bullet}, \;\mathcal{L}_{A,f}=\mathcal{L}_{A,f}^{\bullet}-\gamma_{A,f},\mathcal{L}_{I,f}=\mathcal{L}_{I,f}^{\bullet}-\gamma_{I,f}$, 
$\{\lambda_{A,f}, \lambda_{I,f}\}_{f \in \mathcal{F}}$ are dual variables. 
\label{P:p2}
\end{problem}

\begin{algorithm}[tb]
   \caption{Training a DDE.}
   \label{alg:algs}
\footnotesize
\begin{algorithmic}
   \STATE {\bfseries Input:} Training samples $\mathcal{X}_T$ with Observed factor $\mathcal{F}$, batch size $B$, primal and dual learning rates $\eta_{D},\{\eta_{A,f},\eta_{I,f}\}_{f \in \mathcal{F}}$, Adam hyperparameters  $\beta_1, \beta_2$,  constraint satisfaction margins $\{\gamma_{A,f}, \gamma_{I,f}\}_{f \in \mathcal{F}}$
    \STATE {\bfseries Initialization:} $\theta_s$ Parameters of $\mathcal{E}_s$, initial  dual variable values $\lambda=( \lambda^0_{A,f}, \lambda^0_{I,f})$
    \STATE {\bfseries Output:} $\theta_s^*$ and $\{\lambda^*_{A,f}, \lambda^*_{I,f}\}_{f \in \mathcal{F}}$
   \REPEAT
   \FOR{$i =1$ {\bfseries to} $B$}
    \STATE $\mathcal{L}^i_{D}= \mathcal{L}^{\diamond}_{D}(x_i)$ 
    \FOR {$f \in \mathcal{F}$}
     \STATE $\mathcal{L}^{i}_{A,f}=max \{ \mathcal{L}^{\diamond}_{A,f} (x_i,x'_i)-\gamma_{A,f} ,0 \}$
  \STATE $\mathcal{L}^{i}_{I,f}=max \{ \mathcal{L}^{\diamond}_{I,f} (x_i,x'_i)-\gamma_{I,f} ,0 \}$
    
    \ENDFOR
     \STATE$\mathcal{L}_i= \mathcal{L}^i_{D}+ \sum_{f \in \mathcal{F}} [\lambda_{A} *\mathcal{L}^{i}_{A,f} + \lambda_{I} *\mathcal{L}^{i}_{I,f}]$ 
   \ENDFOR
 \STATE\textbf{Primal step} 
 \STATE     $\theta_s \longleftarrow Adam(\frac{1}{B}\sum_{i=1}^{B}\mathcal{L}_i,\theta_s,\eta_D,\beta_1,\beta_2$) 
      \STATE \textbf{Dual step} 
 \FOR{$f \in \mathcal{F}$}
\STATE $\lambda_{A,f}  \longleftarrow max\{ [\lambda_{A,f}+\eta_{A,f} \frac{1}{B}\sum_{i=1}^{B} \mathcal{L}^{i}_{A,f}],0\}$ 
  \STATE  $\lambda_{I,f}\longleftarrow max\{ [\lambda_{I,f}+\eta_{I,f} \frac{1}{B}\sum_{i=1}^{B} \mathcal{L}^{i}_{I,f}],0\}$ 
   \ENDFOR
   \UNTIL{$\theta_s$ is  converged.}
\end{algorithmic}
\end{algorithm}

Algorithm  ~\ref{alg:algs} shows the primal-dual approach for training the student encoder model. First, main objective  $\mathcal{L}_D$, and constraint losses $\mathcal{L}_{A,f}, \; \mathcal{L}_{I,f}$ are calculated. Then, in the primal step,  the total loss is optimized with respect to encoder parameters  $\theta_s$. In the dual step, dual variables  $\lambda_{A,f}$ and $\lambda_{I,f}$ are increased gradually until their corresponding loss constraints converge to the pre-defined margins.

\subsection{OOD reasoning}
\label{OODReas}
To form OOD reasoners for each factor  $f$, we use the k-means algorithm ~\cite{lloyd1982least} to cluster data in each factor's representative dimensions and approximate the Gaussian mixture model ~\cite{reynolds1992gaussian} based on cluster centers. Test samples with membership probability below a specific threshold $\varsigma_{f}$ are  OOD with respect to factor  $f$.
\section{Analyzing the optimality of solutions}
\label{Optanlz}

For analyzing the optimality of the relaxed DDCO problem, it is required to ensure that the following assumptions regarding the complexity of the student model hold. 

\begin{figure}[t]
\centering
\includegraphics[width=0.8\columnwidth]{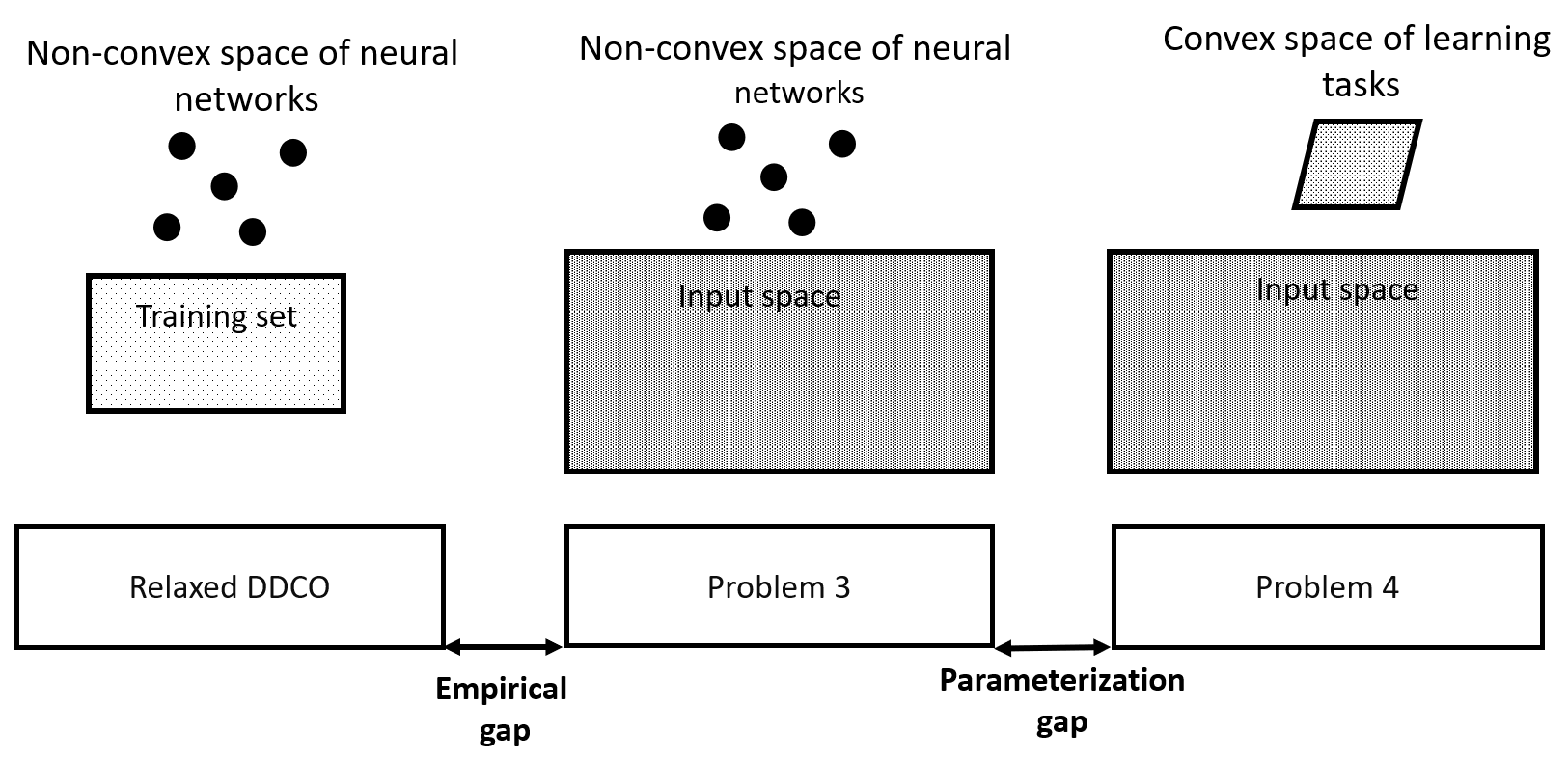}
\caption{Parameterization and empirical gaps.}
\label{fig:Paremp}
\end{figure}

Assumption  ~\ref{as:upcom} limits the complexity of hypothesis space and prevents over-fitting for training data. 
\begin{assumption}[Upper bound on complexity of student hypothesis space]\label{as:upcom}
Consider loss functions   $\mathcal{L}^{\diamond}_{D}, \mathcal{L}^{\diamond}_{A,f}$, and     $\mathcal{L}^{\diamond}_{I,f}$ that are defined  over  distributions     $\mathfrak{D}(x)$ from which i.i.d samples   $x$ and   $x'$ are drawn. 
With a probability of  $1-\delta$, there are functions   $\zeta_{D},\zeta_{A},$ and $ \zeta_{I}$  that bound the distance between real and empirical losses and are  monotonically decreasing  with respect to   $m, \; m_A$  and  $m_I$, respectively:
    \begin{equation}
    \begin{aligned}
    &|E_{x \sim \mathfrak{D}(x)}  [\mathcal{L}^{\diamond}_{D}(x)]-\frac{1}{m} \sum_{i=1}^m   \mathcal{L}^{\diamond}_{D}(x_i) |\le \zeta_{D} \\
    &|E_{(x,x') \sim \mathfrak{D}(x)}  [\mathcal{L}^{\diamond}_{A,f}(x_i,x'_i)]-\frac{1}{m_A} \sum_{i=1}^{m_A}   \mathcal{L}^{\diamond}_{A,f}(x_i,x'_i) |\le \zeta_{A}\\ & |E_{(x,x') \sim \mathfrak{D}(x)}  [\mathcal{L}^{\diamond}_{I,f}(x_i,x'_i)]-\frac{1}{m_I} \sum_{i=1}^{m_I}   \mathcal{L}^{\diamond}_{I,f}(x_i,x'_i) |\le \zeta_{I}
    \end{aligned}
    \end{equation} 
\end{assumption}
Non-convex student hypothesis must be sufficiently complex to cover the output of the convex function it models.
Assumption ~\ref{as:lowcom} states that the non-convex encoder model $\mathcal{E}_s$ can parameterize convex feature extractor function  $C_s$ by $\epsilon_s$ error.
\begin{assumption}[Lower bound on complexity of student hypothesis space]\label{as:lowcom}
Consider the closed convex hull  $\overline{\mathcal{H}_s}$ that contains all the convex hypotheses from the student hypothesis space $\mathcal{H}_s$. Then there exists  $\epsilon_s \geq 0$ and  $\theta_s \in \Theta_s$: 
\begin{equation}
    \forall C_s \in \overline{\mathcal{H}_s}: \;  E_{\mathfrak{D}(x)} [|C_s(x)-\mathcal{E}_s(\theta_s,x)|] \leq \epsilon_s
\end{equation}
\end{assumption}
Based on assumptions ~\ref{as:upcom} and ~\ref{as:lowcom} for a student model,  non-optimality for solutions of  Problem ~\ref{P:p2} stem from empirical and parameterization gaps ~\citep{chamon2022constrained}. Empirical gap occurs when a student encoder is trained on training samples instead of the entire input space, while the parameterization gap arises when a non-convex encoder model learns convex tasks such as feature extraction.
 Problem ~\ref{P:p2} is redefined over input space (Problem ~3   of Table ~\ref{TB:OPTPD}) and convex function space  (Problem ~4   of Table ~\ref{TB:OPTPD}) to analyze empirical and parametrization gaps, respectively. 
Figure ~\ref{fig:Paremp} shows the parameterization, empirical gaps, and corresponding problems. 

Table ~\ref{TB:OPTPD} contains the required primal and dual problems for analyzing parameterization and empirical gaps. Due to the complexity of solving a constrained problem for optimizers, we define a dual unconstrained problem based on the Lagrangian ~\citep{boyd2004convex} for all primal problems in Table~\ref{TB:OPTPD}. $\lambda$ and its variants are dual variables. Table~\ref{TB:conslosses} expand definitions for used losses in Table ~\ref{TB:OPTPD}. In this table  $\tilde{\mathcal{L}}^{\diamond}_{D}(x), \; \tilde{\mathcal{L}}^{\diamond}_{A,f}(x,x'), \; \tilde{\mathcal{L}}^{\diamond}_{I,f}(x,x')$ are defined by substituting non-convex encoders   $\mathcal{E}_s$  and   $\mathcal{E}_{\tau}$  with convex  feature extractors $C_s \in \overline{\mathcal{H}}_s$  and   $C_\tau \in \overline{\mathcal{H}_{\tau}}$ in   $\mathcal{L}^{\diamond}_D, \; \mathcal{L}^{\diamond}_A$  and   $\mathcal{L}^{\diamond}_I$, respectively.

The empirical gap, denoted by $\vert\hat{d}^*-d^*\vert$, represents the gap between the optimal solutions of dual problems defined over training and input spaces. The parameterization gap is the distance between optimal values of the disentanglement distillation problem when it is defined over convex function space  $\overline{\mathcal{H}}_s$ and non-convex hypothesis space  $\mathcal{H}_s$  ($\vert \tilde{p}^*-\hat{d}^*\vert$).

Problem~5 in Table~\ref{TB:OPTPD} is a perturbed version of Problem~3 from this table and is used in Proposition~\ref{th:bounds} to derive a parameterization gap by connecting the optimal solutions of the defined problems for convex functions and non-convex hypotheses.

\begin{table*}[t]
\caption{Required primal and dual optimization problems for analyzing the optimality of solutions obtained by DDE.}
\centering
\resizebox{0.7\textwidth}{!}{%
\begin{tabular}{|c|c|c|c|}
\hline
No. & \begin{tabular}[c]{@{}c@{}}  Conditions      \end{tabular}                                                                                              & Primal form & Dual form \\ \hline

$3$  & \begin{tabular}[l]{@{}c@{}}  Non-convex  hypothesis\\ over  input space\end{tabular} &     $ \begin{aligned}
        & \hat{p}^* \triangleq & min_{\theta_s \in \Theta_s} \hat{\mathcal{L}}_D^{\bullet}(\theta_s)\\
        &subject.to &\hat{\mathcal{L}}_{A,f}^{\bullet}(\theta_s) \leq \gamma_{A,f} \; (\forall f \in \mathcal{F}) \\
        &  &\hat{\mathcal{L}}_{I,f}^{\bullet}(\theta_s) \leq \gamma_{I,f} \; (\forall f \in \mathcal{F})
    \end{aligned}$     &   $\begin{aligned}
        &\hat{d}^* \triangleq max_{\{\hat{\lambda}_{A,f},\hat{\lambda}_{I,f}\}_{f \in \mathcal{F}}} min_{\theta_s \in \Theta_s} \\ &\hat{\mathcal{L}}_{D}+ \sum_{f \in \mathcal{F}}
         \hat{\lambda}_{A,f} *\hat{\mathcal{L}}_{A,f} +
         \hat{\lambda}_{I,f} *\hat{\mathcal{L}}_{I,f}
    \end{aligned}$      \\ \hline

$4$  & \begin{tabular}[l]{@{}c@{}}  Convex  function\\ over  input space\end{tabular} &     $\begin{aligned}
          &\tilde{p^*} \triangleq & min_{C_s \in  \overline{\mathcal{H}_s}}\tilde{\mathcal{L}}_{D}^{\bullet}(C_s)\\
        &{subject.to}& {\; \tilde{\mathcal{L}}^{\bullet}_{A,f}(C_s) \leq \gamma_{A,f} \; (\forall f \in \mathcal{F})} \\
        &  &{\tilde{\mathcal{L}}^{\bullet}_{I,f}(C_s) \leq \gamma_{I,f} \; (\forall f \in \mathcal{F})}
    \end{aligned}$     &  $\begin{aligned}
        &\tilde{d}^* \triangleq max_{\{\tilde{\lambda}_{A,f}, \tilde{\lambda}_{I,f}\}_{f \in \mathcal{F}}} min_{C_s \in \overline{\mathcal{H}}_s} \\
        &\tilde{\mathcal{L}}_{D}+ \sum_{f \in \mathcal{F}}
         \tilde{\lambda}_{A,f} *\tilde{\mathcal{L}}_{A,f} +
         \tilde{\lambda}_{I,f} *\tilde{\mathcal{L}}_{I,f}
    \end{aligned}$       \\ \hline

$5$  & \begin{tabular}[l]{@{}c@{}} Perturbed problem \\ defined with convex  \\ function over  input space\end{tabular} &     
         $\begin{aligned}
  & \tilde{p}_{\varkappa }^* \triangleq &
  min_{C_s\in \overline{\mathcal{H}}_s} \tilde{\mathcal{L}}_{D}^{\bullet}(C_s)\\
        & subject \; to: & \tilde{\mathcal{L}}_{A,f}^{\bullet}(C_s) \leq \gamma_{A,f} -\kappa_A *{\epsilon}\; (\forall f \in \mathcal{F}) \\
        &  &\tilde{\mathcal{L}}_{I,f}^{\bullet}(C_s) \leq \gamma_{I,f}  -\kappa_I *{\epsilon} \; (\forall f \in \mathcal{F})  
    \end{aligned}$     &   $\begin{aligned}
       &\tilde{d}_{\varkappa} \triangleq  max_{\{\tilde{\lambda}_{A,f,\varkappa}, \tilde{\lambda}_{I,f,\varkappa}\}_{f \in \mathcal{F}}} min_{C_s \in \overline{\mathcal{H}}_s} \\ 
       &\tilde{\mathcal{L}}_{D,\varkappa}+
         \tilde{\lambda}_{A,f,\varkappa} *\tilde{\mathcal{L}}_{A,f,\varkappa} +
         \tilde{\lambda}_{I,f,\varkappa} *\tilde{\mathcal{L}}_{I,f,\varkappa} 
    \end{aligned}$   
    \\ \hline
\end{tabular}
}

\label{TB:OPTPD}
\end{table*}


\begin{table}[]
\caption{Loss functions definitions.}

\centering
\resizebox{0.8\columnwidth}{!}{%
\begin{tabular}{|ll|}
\hline
\multicolumn{1}{|c|}{\textbf{Primal loss}}                                    & \multicolumn{1}{c|}{\textbf{Description}} \\ \hline

\multicolumn{1}{|l|}{$\hat{\mathcal{L}}_{D}^{\bullet}(\theta_s)=  E_{x \sim \mathfrak{D}(x)} \mathcal{L}^{\diamond}_{D}(x)$}                                                                                                              & True distillation loss                                                                                                       \\ \hline
\multicolumn{1}{|l|}{$\hat{\mathcal{L}}_{A,f}^{\bullet}(\theta_s) =  E_{(x,x') \sim \mathfrak{D}(x)}\mathcal{L}^{\diamond}_{A,f} (x,x')$}                            & True adaptation loss                                                                                                       \\ \hline

\multicolumn{1}{|l|}{$\hat{\mathcal{L}}_{I,f}^{\bullet}(\theta_s) =  E_{(x,x') \sim \mathfrak{D}(x)}{\mathcal{L}}^{\diamond}_{I,f} (x,x')$}                                                                                                                                                                                                       & True isolation loss                                                                                                       \\ \hline

\multicolumn{1}{|l|}{$\tilde{\mathcal{L}}_{D}^{\bullet}(C_s)\triangleq  E_{x \sim \mathfrak{D}(x)} \tilde{\mathcal{L}}^{\diamond}_{D}(x)$}                                                                                                                                                                                &\begin{tabular}[c]{@{}l@{}} True distillation loss  \\defined over function space         \end{tabular}                                                                                                         \\ \hline

\multicolumn{1}{|l|}{$\tilde{\mathcal{L}}_{A,f}^{\bullet}(C_s) =  E_{(x,x') \sim \mathfrak{D}(x)}\tilde{\mathcal{L}}^{\diamond}_{A,f} (x,x')$}                                                                                                                                                                                                        &  \begin{tabular}[c]{@{}l@{}} True adaptation loss  \\defined over function space         \end{tabular}                                                                                                       \\ \hline

\multicolumn{1}{|l|}{$\tilde{\mathcal{L}}_{I,f}^{\bullet}(C_s) =  E_{(x,x') \sim \mathfrak{D}(x)}\tilde{\mathcal{L}}^{\diamond}_{I,f} (x,x')$}                                                                                                                                                                                                        &  \begin{tabular}[c]{@{}l@{}} True isolation loss  \\defined over function space         \end{tabular}                                                                                             \\ \hline
\multicolumn{2}{|c|}{\textbf{Dual loss}}                                                                                                                                                                                   \\ \hline

\multicolumn{2}{|l|}%
{$\hat{\mathcal{L}_{D}}=\hat{\mathcal{L}_{D}^{\bullet}}, \hat{\mathcal{L}}_{A,f}=\hat{\mathcal{L}}_{A,f}^{\bullet}-\gamma_{A,f},\hat{\mathcal{L}}_{I,f}=\hat{\mathcal{L}}_{I,f}^{\bullet}-\gamma_{I,f}
 $}    \\ \hline
\multicolumn{2}{|l|}{$\tilde{\mathcal{L}_{D}}=\tilde{\mathcal{L}_{D}^{\bullet}} , \tilde{\mathcal{L}}_{A,f}=\tilde{\mathcal{L}}_{A,f}^{\bullet}-\gamma_{A,f}, \tilde{\mathcal{L}}_{I,f}=\tilde{\mathcal{L}}_{I,f}^{\bullet}-\gamma_{I,f} $}    \\ \hline
\multicolumn{2}{|l|}{\begin{tabular}[c]{@{}l@{}} 
 $\tilde{\mathcal{L}}_{D,\varkappa}=\tilde{\mathcal{L}_{D}^{\bullet}} ,\tilde{\mathcal{L}}_{A,f,\varkappa}=\tilde{\mathcal{L}}_{A,f}^{\bullet}-\gamma_{A,f}+\kappa_A *{\epsilon}$, \\ $\tilde{\mathcal{L}}_{I,f,\varkappa}=\tilde{\mathcal{L}}_{I,f}^{\bullet}-\gamma_{I,f}+\kappa_I *{\epsilon}$        \end{tabular}}
 \\ \hline
\end{tabular}%
}
\label{TB:conslosses}
\end{table}

For the validity of the empirical and parameterization gap definitions, strong duality must hold for problems ~\ref{P:p2} and Problem~3 of Table~\ref{TB:OPTPD}. Strong duality holds for these problems under adapted conditions from ~\citep{chamon2022constrained} and also feasibility assumptions for problems ~\ref{P:p2} and  Problem ~3 from Table ~\ref{TB:OPTPD}. Feasibility assumptions ensure there is at least one valid solution for these problems (refer to Appendix \ref{A:FAss} for  formal definitions of feasibility assumptions). 

Theorem ~1 from ~\citep{chamon2022constrained} is defined initially for supervised settings. However, it can also be adapted to a match-pairing setting to analyze the empirical and parameterization gaps. Data partitions from section ~\ref{Datapart} can be seen as implicit labels, where a training sample is $\{(x_i,y_i)\}$, with $y_i \in \mathcal{Y}$ being the index of the group that contains a training sample $x_i$. As the following conditions hold in this setting, Theorem  ~1 can be used.
\begin{enumerate}
    \item Set  $\mathcal{Y}$ is finite:
    Set   $\mathcal{Y}$ is finite as the number of partitions is finite and equal to  K. 
    \item Non-atomicity of drawn  random variables from probability distribution $\mathfrak{D}(x)$: 
     Non-atomicity means samples derived from the distribution $\mathfrak{D}(x)$ are not identical. Continuous distributions are non-atomic. This condition holds as $\mathfrak{D}(x)$ is a fixed, unknown, and continuous distribution. Also, it is assumed that samples are independent and identically distributed.
     \item  $\overline{\mathcal{H}_s}$ is decomposable: The teacher encoder model is trained using a prior Gaussian distribution that assumes a Euclidean latent space. Since the student model imitates the teacher's latent space, the    $\overline{\mathcal{H}_s}$ that the encoder model parameterizes is also a Euclidean space. As Euclidean, and in general, Lebesgue spaces ~\citep{castillo2016introductory}, are decomposable,   $\overline{\mathcal{H}_s}$ is decomposable ~\citep{kalatzis2020variational}.
\end{enumerate}
Proposition ~\ref{th:bounds} provides upper bounds for parameterization, empirical gaps, and the expectation of loss functions.
\begin{proposition}[From Theorem 1 in ~\citep{chamon2022constrained}]
Suppose conditions 1-3  hold.
   $\lambda^*=\{\lambda_{A,f}^*,\lambda_{I,f}^*\}_{f \in \mathcal{F}}$ is the optimum dual variable for Problem~\ref{P:p2}. Under Assumptions ~\ref{as:LIPBOUND}, ~\ref{as:upcom}, ~\ref{as:lowcom} and feasibility assumptions (Appendix \ref{A:FAss}), there exists an optimal prime value $\theta^*$ for Problem~\ref{P:p2} such that with probability $1-(3*2*\vert\mathcal{F}\vert+2) *\delta$:
\begin{equation}
    \begin{aligned}
    &\vert\tilde{p}^*-\hat{d}^*\vert \leq (1+\Vert\tilde{\lambda}^*_{\varkappa}\Vert_1) (\overline{\kappa} * \epsilon)
    \label{eq:pargap}
    \end{aligned}
\end{equation}
    \begin{equation}
        \begin{aligned}
            &\vert\hat{d}^*-d^*\vert \leq (1+max \{\Vert\lambda^*||_1,||\hat{\lambda}^*\Vert_1 \}) *\overline{\zeta}
        \end{aligned}
        \label{eq:empgap}
    \end{equation} 
   \begin{equation}
    \begin{aligned} 
    &E_{x \sim \mathfrak{D}(x)}[\mathcal{L}^{\diamond}_{D}(x)] \leq \zeta_{D} \\
    &E_{(x,x') \sim \mathfrak{D}(x)}[\mathcal{L}^{\diamond}_{A,f} (x,x')] \leq \gamma_{A,f} + \zeta_A\\
    &E_{(x,x') \sim \mathfrak{D}(x)} [\mathcal{L}^{\diamond}_{I,f} (x,x')] \leq \gamma_{I,f} + \zeta_I\\
    \end{aligned}
    \label{eq:empbound}
    \end{equation}
    Here,  $\overline{\zeta}=max (\zeta_A,\zeta_I)$, $\overline{\kappa} = max(\kappa_D,\;\kappa_A,\;\kappa_I)$, and  $\epsilon=max(\epsilon_s,\epsilon_{\tau})$.  $\kappa_D$, $\kappa_A$, and  $\kappa_I$ are Lipschitz constants for distillation, adaptation, and isolation losses, respectively.  $\lambda^*,\hat{\lambda}^*$ and   $\tilde{\lambda}_{\varkappa}^*$ are optimal dual variables for dual Problems~\ref{P:p2}, 3 of Table ~\ref{TB:OPTPD} and 5 of Table ~\ref{TB:OPTPD}, respectively.
\label{th:bounds}
\end{proposition}
Equation ~\ref{eq:pargap} indicates that the parameterization gap depends on loss function sensitivity to change in output of student encoder ($\overline{\kappa}$), student and teacher model ability to learn a given task ($\epsilon$), and perturbed constraint satisfaction. Equation Equation~\ref{eq:empgap}   relates the empirical gap to constraint satisfaction when using training data, input space, and model complexity. Equation ~\ref{eq:empbound}  shows that the expectation of each loss function is limited by its complexity and preset margin of constraint satisfaction.
In practice, it is impossible to calculate the parameterization and empirical gaps due to their reliance on the optimal value of dual variables of abstract constrained optimization problems defined in convex functional spaces or with infinite data. 

To upper bound the expectation of each loss function
Rademacher complexity (RC) ~\citep{mohri2018foundations} is used. RC measures the difference between true and empirical losses defined over input and training data. 
The Lipschitz coefficient of the student encoder model can control RC by measuring the encoder's sensitivity to input data changes. This coefficient is calculated by the operations of its layers. Convolution layers can be represented as linear operators  $\textsc{OP}(\mathcal{R})$ ~\citep{lecun2015deep}, and expressed as  $\vert\mathcal{R}\vert$-ly block circulant matrices (in these matrices the elements of each row are the shifted variation of the previous row)
 ~\citep{long2019generalization}, with  $\vert\mathcal{R}\vert$ being the size of the convolution kernel. For a linear layer, the operator is identical to a matrix that indicates the layer operation. The Lipschitz coefficient of the student encoder is determined by calculating the singular values of the linear operators of its layers as follows ~\citep{sedghi2018singular}.

\begin{definition}[Lipschitz coefficient of  student encoder] \label{def:LipNet}
Consider a student encoder with $L$ layers including convolution  $\{\mathcal{R}_1,..., \mathcal{R}_{L_\mathcal{R}}\}$ and linear layers  $\{\mathcal{Q}_1, ..., \mathcal{Q}_{L_{\mathcal{Q}}} \}$. Suppose the weight initializations in convolution and linear layers are specified as   $\forall i \in L_{\mathcal{R}}: \mathcal{R}^0_i$ and $\forall i \in  L_{\mathcal{Q}}: \mathcal{Q}^0_i$ and they are bounded by   $1+\nu$ ($\forall i \in L_{\mathcal{R}}: \; ||\textsc{OP}(\mathcal{R}^0_i)||_2 \leq 1+\nu$, $\forall i \in L_{\mathcal{Q}}: \Vert\textsc{OP}(\mathcal{Q}^0_i)\Vert_2 \leq 1+\nu$).  In addition, the distance between learned and initial weights is bounded  ($ \sum _{i\in L_\mathcal{R}}\Delta^{\mathcal{R}}_i +  \sum _{i\in L_\mathcal{Q}}\Delta^{\mathcal{Q}} _i\leq \Delta_{op}$) where  $\sum_{i \in L_{\mathcal{R}}}\vert\textsc{OP}(\mathcal{R}_i)-\textsc{OP}(\mathcal{R}_i^0)\vert\leq \Delta^{\mathcal{R}}_i$ and  $\sum_{i \in L_{\mathcal{Q}}}\vert \textsc{OP}(\mathcal{Q}_i)-\textsc{OP}(\mathcal{Q}^0_i)\vert \leq \Delta^{\mathcal{Q}}_i $.
Suppose $m$ samples with flattened Euclidean norm less than  $\chi$  ($\forall x \in \mathcal{X}_{T}:\Vert vec(x)\Vert_2 \leq \chi$). Also, consider $\kappa$ as the Lipschitz coefficient of the loss function. Then, the network Lipschitz coefficient is defined as follows:
    \begin{equation}
        \begin{aligned}
           &\kappa_{\theta}= \chi*\kappa*\Delta_{OP}*(1+\nu+\frac{\Delta_{OP}}{L})^L
        \end{aligned}
            \label{eq:LipNets}
    \end{equation}
    \label{df:NNLips}
\end{definition}
Proposition ~\ref{prp:gbtt} provides an upper bound over the expectation of the distillation loss function. For other losses, we follow the same steps (refer to Appendixes ~\ref{sbc:RCas} and ~\ref{app:RC}).
\begin{proposition}
[Bound over expectation of loss (from Theorem 2 of ~\citep{foster2019hypothesis})]
	Consider a  $\kappa_D\omega$-stable student hypothesis space   $\mathcal{H}_s$  with CV-stability. The stability of the hypothesis means that a slight change in its training sample does not lead to drastic changes in its output (refer to Appendix ~\ref{sbc:RCas} for required assumptions and Proposition ~\ref{lm:stb} for formal definition). CV-stability means that the loss obtained by the student hypothesis does not drastically change by substituting one sample with another during training ~\citep{foster2019hypothesis} (refer to  Assumption ~\ref{as:stbstd} for formal definition). Also, the Lipschitz coefficient and bound over a range of loss are $\kappa_D$ and $B_D$, respectively.
 Then, for any $\delta \geq 0$ with a probability of  $1-\delta$ and a student model  $\mathcal{E}_s \in \mathcal{H}_s$, the gap between true and trained losses is defined as follows:
\begin{equation}
\begin{aligned}
    &E_{x \sim \mathfrak{D}(x)} \mathcal{L}^{\diamond}_{D}(x)-\frac{1}{m} \sum_{i=1}^{m} \mathcal{L}^{\diamond}_{D}(x)\\
    & \leq 2* R_m^\diamond (\mathcal{L}^{\diamond}_{D}(x))+ (B_D+2 \kappa_D \omega m)*\sqrt{\frac{1}{2m}ln \frac{1}{\delta}}
\end{aligned}	
\label{eq:eq15}
\end{equation}
 \label{prp:gbtt}
\end{proposition}
Since the loss function is Lipschitz parameterized 
(refer to Assumption ~\ref{as:parlips}), based on Talagrand's lemma ~\citep{mohri2016learning},  the upper bound for $R_m^\diamond (\mathcal{L}^{\diamond}_{D}(x))$ is calculated by the empirical RC ($E_{\mathcal{X}}[R_m^{\diamond}(en_s)]$). Then, based on the Dudley entropy integral ~\citep{bartlett2013theoretical}:
\begin{equation}
    E_{\mathcal{X}}[R_m^{\diamond}(\mathcal{E}_s)]   \leq \kappa_{\theta} \sqrt{\frac{8.7*d}{m}}
\end{equation}
  Then, by replacing $R_m^\diamond (\mathcal{L}^{\diamond}_{D}(x))$ with $E_{\mathcal{X}}[R_m^{\diamond}(\mathcal{E}_s)]$ in Equation ~\ref{eq:eq15} and substituting the Lipschitz coefficient of $\mathcal{E}_s$ with Equation~\ref{eq:LipNets} in the Dudley theorem Equation ~\ref{eq:empgapfin2} is derived.
\begin{equation}
    \begin{aligned} 
        & \zeta_{D}= 
        2*\chi*\kappa_D*\Delta_{OP}*(1+\nu+\frac{\Delta_{OP}}{L})^L \sqrt{\frac{8.7*d}{m}} \\
        &+(B_D+2 \kappa_D \omega m)*\sqrt{\frac{1}{2m}ln \frac{1}{\delta}}
    \end{aligned}
    \label{eq:empgapfin2}
\end{equation}


\section{Implementation and Evaluation}
We evaluate our approach by applying it to the CARLA dataset ~\citep{Dosovitskiy2017}. We used a desktop computer with Geforce RTX   $3080$  and  $64 \; GB$ memory to train the teacher and student models. We use   WDLVAE  ~\citep{rahiminasab2022out}  as a teacher model (refer to Appendix ~\ref{scb:Techach} for architecture details) as it is designed for OOD reasoning for multi-label data and has partially disentangled latent space. In designing the student architecture,  we remove   $10\%-90\%$ (compression rate  $r \in [0.1,0.9]$) of the neurons from each layer of the teacher encoder, augment batch normalization and convolution layers and set the number of epochs to   $50$. We use the same data and partitions presented in the WDLVAE for a fair comparison between the teacher and student models. The selected generative factors are rain (\textsc{R}) and background (\textsc{BK}), and we obtained data partitions by combining different values for these factors. We had 3000 training and  600 calibration samples, with  2592 and  1296 test samples to evaluate the rain and background reasoners, respectively. Details about partitions are mentioned in Appendix ~\ref{App:Datgen}. For both teacher and student, the representative dimensions for rain and background factors are set to  3 and 6, respectively. 
\begin{figure}[t!]
    \centering
    \includegraphics[width=0.6\columnwidth]{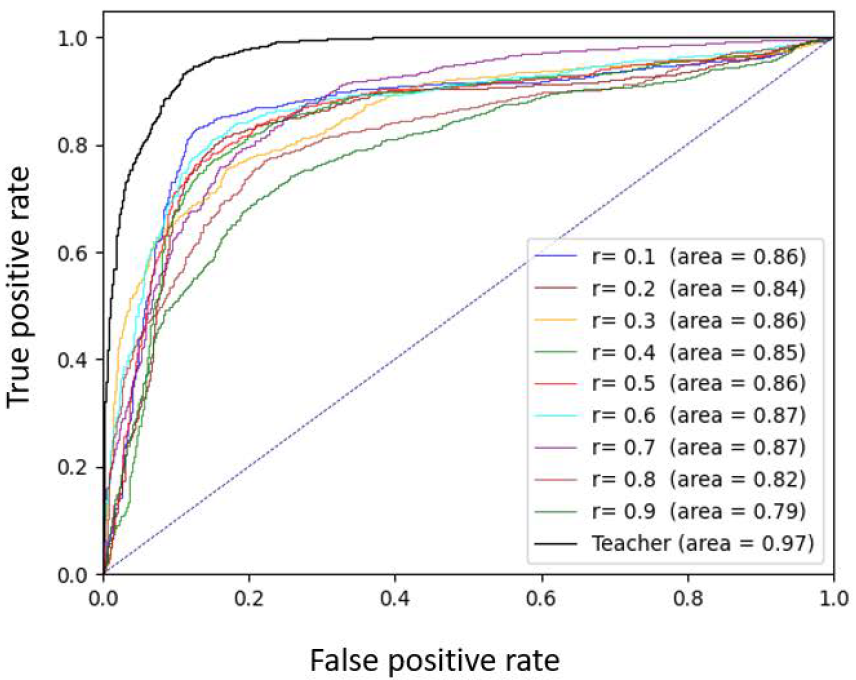}
    \caption{AUROC curve for rain reasoner.}
    \label{fig:AUROCrain}
\end{figure}

\begin{figure}[t!]
    \centering
    \includegraphics[width=0.6\columnwidth]{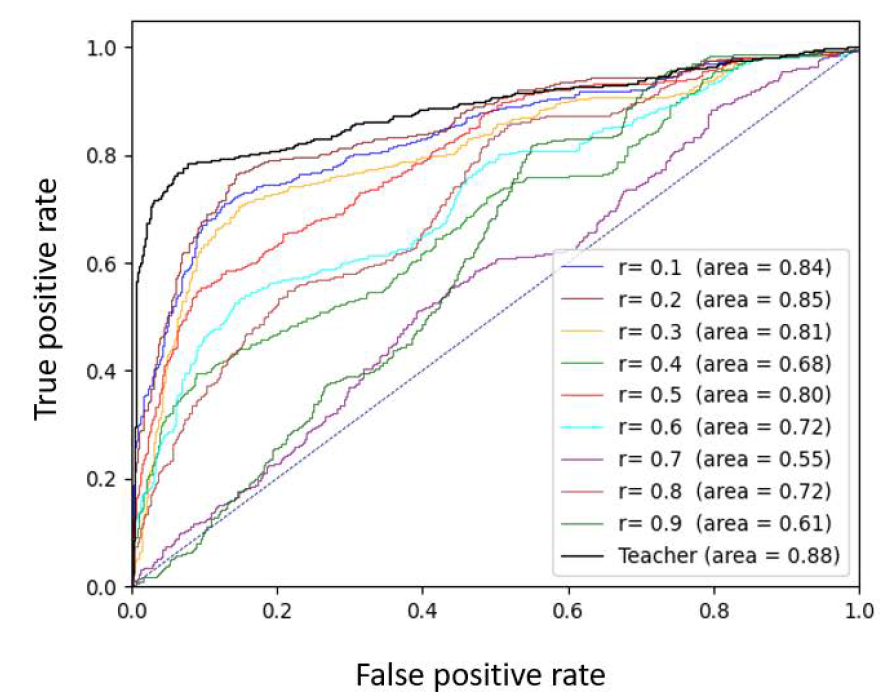}
    \caption{AUROC curve for  background reasoner.}
    \label{fig:AUROBack}
\end{figure}

\begin{figure}[t!]
\centering
\includegraphics[width=0.6\columnwidth]{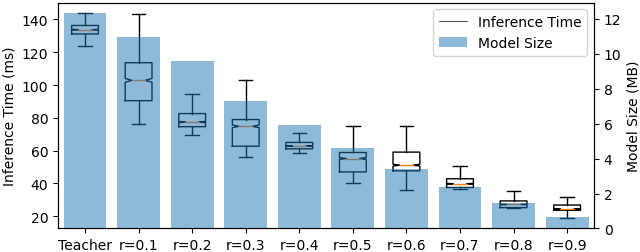} 
\caption{Inference time and model size of compressed student models vs. teacher model.}
\label{fig:times}
\end{figure}


The closest approaches to our study are ~\citep{robey2021model,zhang2022towards}. However, we did not compare our approach to them as they solve DG problems rather than OOD reasoning, in which information regarding OOD data may be available during training. Also, they are designed for single-label data and are resource-intensive as the number of required models grows linearly with respect to the number of content elements (in the OOD problem, content elements can be seen as generative factors).

We evaluate our approach based on OOD reasoning performance, required model size, and test inference time.

Figures ~\ref{fig:AUROCrain} and ~\ref{fig:AUROBack} show that the teacher model has AUROCs of $97\%$ and $88\%$ for rain and background factors. Despite a slight decrease in AUROC at the start of compression for student models, our approach maintains AUROC stability until  $r=0.7$ and $r=0.5$ for rain and background reasoners, respectively. These numbers indicate that disentanglement constraints are enforced during compression. So, we can compress the model $50\%$ while the performance is preserved around $86\%$ $80\%$ for rain and background reasoners, respectively.

We ran the models on a Jetson Nano with  $4$ CPU cores to measure memory usage and inference time. CPU execution was chosen due to the need for timely processing in a real CPS where another ML model may occupy the GPU (refer to Appendix ~\ref{scb:TM} for details). Figure~\ref{fig:times} shows that increasing compression rates decreases the model size and inference time. For compression rate $50 \%$, which has proper OOD performance,   the model size and average inference time are $4.37 \; MB$ and $ 54.33 \; ms$  compared to model size $12.4 \; MB$  and average inference time $ 131.83\; ms$ for the teacher model.

In  Appendix ~\ref{scb:appres}, we also show that the disentanglement constraints are satisfied, and RC is well-defined for distillation and disentanglement loss functions.


\section{Conclusion}
This paper presents a DDE framework that decreases OOD reasoner size while preserving its latent space disentanglement. DDE is trained as a constrained optimization problem. The optimality of the obtained solutions for this problem is analyzed based on parameterization and empirical gaps. This approach is evaluated with the CARLA dataset on Jetsen Nano. In the future, we plan to extend this study to other compression methods, such as pruning, and consider the role of temporal dependency in defining disentanglement.



\bibliography{uai2024-template}

@phdthesis{reynolds1992gaussian,
    author = {Reynolds, Douglas Alan},
    title = {A Gaussian Mixture Modeling Approach to Text-Independent Speaker Identification},
    school = {Georgia Institute of Technology},
    year = {1992},
    type = {Ph.D. diss.},
    month = aug,
    address = {Atlanta, GA, USA}
}

@book{boyd2004convex,
    title = {Convex Optimization},
    author = {Boyd, Stephen and Vandenberghe, Lieven},
    year = {2004},
    publisher = {Cambridge University Press},
    address = {Cambridge, UK}
}

@book{mohri2018foundations,
    title = {Foundations of Machine Learning},
    author = {Mohri, Mehryar and Rostamizadeh, Afshin and Talwalkar, Ameet},
    year = {2018},
    publisher = {MIT Press},
    address = {Cambridge, MA, USA}
}

@book{castillo2016introductory,
    title = {An Introductory Course in Lebesgue Spaces},
    author = {Castillo, Ren{\'e} Erl{\'\i}n and Rafeiro, Humberto},
    year = {2016},
    publisher = {Springer Cham},
    address = {Cham, Switzerland}
}

@book{wild1947arctangent,
  title={The Arctangent Function and Its Application to the Computation of Pi},
  author={Wild, R. E. and Chittenden, E. W.},
  url={https://books.google.com.sg/books?id=v1KsYgEACAAJ},
  year={1947},
  publisher={University of Iowa}
}

@book{goodfellow2016deep,
  title={Deep learning},
  author={Goodfellow, Ian and Bengio, Yoshua and Courville, Aaron},
  year={2016},
  publisher={MIT press}
}

@incollection{zniyed2022structured,
    title = {Structured tensor train decomposition for speeding up kernel-based learning},
    author = {Zniyed, Yassine and Karmouda, Ouafae and Boyer, R{\'e}my and Boulanger, J{\'e}r{\'e}mie and de Almeida, Andr{\'e} LF and Favier, G{\'e}rard},
    booktitle = {Tensors for Data Processing},
    pages = {537--563},
    year = {2022},
    publisher = {Elsevier},
    doi = {10.1016/B978-0-12-824447-0.00020-0}
}

@article{lloyd1982least,
    title = {Least Squares Quantization in PCM},
    author = {Lloyd, Stuart},
    journal = {IEEE Transactions on Information Theory},
    volume = {28},
    number = {2},
    pages = {129--137},
    year = {1982},
    publisher = {IEEE},
    doi = {10.1109/TIT.1982.1056489}
}

@article{lin1991divergence,
    title = {Divergence Measures Based on the Shannon Entropy},
    author = {Lin, Jianhua},
    journal = {IEEE Transactions on Information theory},
    volume = {37},
    number = {1},
    pages = {145--151},
    year = {1991},
    publisher = {IEEE},
    doi = {10.1109/18.61115}
}

@article{chamon2022constrained,
    title = {Constrained Learning With Non-Convex Losses},
    author = {Chamon, Luiz F. O. and Paternain, Santiago and Calvo-Fullana, Miguel and Ribeiro, Alejandro},
    journal = {IEEE Transactions on Information Theory},
    volume = {69},
    number = {3},
    pages = {1739-1760},
    year = {2022},
    publisher = {IEEE},
    doi = {10.1109/TIT.2022.3187948}
}

@article{wang2022disentangled,
    title = {Disentangled Representation for Cross-Domain Medical Image Segmentation},
    author = {Wang, Jie and Zhong, Chaoliang and Feng, Cheng and Zhang, Ying and Sun, Jun and Yokota, Yasuto},
    journal = {IEEE Transactions on Instrumentation and Measurement},
    volume = {72},
    pages = {1--15},
    year = {2023},
    publisher = {IEEE},
    doi = {10.1109/TIM.2022.3221131}
}

@article{mohri2016learning,
    title = {Learning Algorithms for Second-Price Auctions with Reserve},
    author = {Mohri, Mehryar and Medina, Andrés Muñoz},
    journal = {Journal of Machine Learning Research},
    volume = {17},
    number = {1},
    pages = {2632--2656},
    year = {2016},
    publisher = {JMLR}
}

@article{lecun2015deep,
    title = {Deep learning},
    author = {LeCun, Yann and Bengio, Yoshua and Hinton, Geoffrey},
    journal = {Nature},
    volume = {521},
    number = {7553},
    pages = {436--444},
    year = {2015},
    publisher = {Nature Research},
    doi = {10.1038/nature14539}
}

@article{hoecker1996svd,
    title = {SVD approach to data unfolding},
    author = {H\"{o}cker, Andreas and Kartvelishvili, Vakhtang},
    journal = {Nuclear Instruments and Methods in Physics Research Section A: Accelerators, Spectrometers, Detectors and Associated Equipment},
    volume = {372},
    number = {3},
    pages = {469--481},
    year = {1996},
    publisher = {Elsevier},
    doi = {10.1016/0168-9002(95)01478-0}
}

@article{gou2021knowledge,
  title={Knowledge distillation: A survey},
  author={Gou, Jianping and Yu, Baosheng and Maybank, Stephen J and Tao, Dacheng},
  journal={International Journal of Computer Vision},
  volume={129},
  pages={1789--1819},
  year={2021},
  publisher={Springer}
}

@article{zhou2021domain,
  title={Domain generalization in vision: A survey},
  author={Zhou, Kaiyang and Liu, Ziwei and Qiao, Yu and Xiang, Tao and Loy, Chen Change},
  journal={arXiv preprint arXiv:2103.02503},
  year={2021}
}

@article{shu2019weakly,
  title={Weakly supervised disentanglement with guarantees},
  author={Shu, Rui and Chen, Yining and Kumar, Abhishek and Ermon, Stefano and Poole, Ben},
  journal={arXiv preprint arXiv:1910.09772},
  year={2019}
}

@article{ramakrishna2022efficient,
  title={Efficient out-of-distribution detection using latent space of $\beta$-vae for cyber-physical systems},
  author={Ramakrishna, Shreyas and Rahiminasab, Zahra and Karsai, Gabor and Easwaran, Arvind and Dubey, Abhishek},
  journal={ACM Transactions on Cyber-Physical Systems (TCPS)},
  volume={6},
  number={2},
  pages={1--34},
  year={2022},
  publisher={ACM New York, NY}
}

@inproceedings{rahiminasab2022out,
  title={Out of Distribution Reasoning by Weakly-Supervised Disentangled Logic Variational Autoencoder},
  author={Rahiminasab, Zahra and Yuhas, Michael and Easwaran, Arvind},
  booktitle={2022 6th International Conference on System Reliability and Safety (ICSRS)},
  pages={169--178},
  year={2022},
  organization={IEEE}
}

@article{nalisnick2019detecting,
  title={Detecting out-of-distribution inputs to deep generative models using typicality},
  author={Nalisnick, Eric and Matsukawa, Akihiro and Teh, Yee Whye and Lakshminarayanan, Balaji},
  journal={arXiv preprint arXiv:1906.02994},
  year={2019}
}

@inproceedings{plumerault2019controlling,
  title={Controlling generative models with continuous factors of variations},
  author={Plumerault, Antoine and Le Borgne, Herv{\'e} and Hudelot, C{\'e}line},
  booktitle={International Conference on Learning Representations},
  year={2019}
}

@inproceedings{xiang2021disunknown,
    title = {DisUnknown: Distilling Unknown Factors for Disentanglement Learning},
    author = {Xiang, Sitao and Gu, Yuming and Xiang, Pengda and Chai, Menglei and Li, Hao and Zhao, Yajie and He, Mingming},
    booktitle = {2021 IEEE/CVF International Conference on Computer Vision (ICCV)},
    pages = {14810--14819},
    year = {2021},
    doi = {10.1109/ICCV48922.2021.01454}
}

@inproceedings{zhang2022towards,
    title = {Towards Principled Disentanglement for Domain Generalization},
    author = {Zhang, Hanlin and Zhang, Yi-Fan and Liu, Weiyang and Weller, Adrian and Sch{\"o}lkopf, Bernhard and Xing, Eric P},
    booktitle = {2022 IEEE/CVF Conference on Computer Vision and Pattern Recognition (CVPR)},
    pages = {8024--8034},
    year = {2022},
    doi = {10.1109/CVPR52688.2022.00786}
}

@inproceedings{yang2022factorizing,
    title = {Factorizing Knowledge in Neural Networks},
    author = {Yang, Xingyi and Ye, Jingwen and Wang, Xinchao},
    booktitle = {Computer Vision -- ECCV 2022},
    pages = {73--91},
    year = {2022},
    doi = {10.1007/978-3-031-19830-4_5}
}

@inproceedings{cha2022domain,
    title = {Domain Generalization by Mutual-Information Regularization with Pre-trained Models},
    author = {Cha, Junbum and Lee, Kyungjae and Park, Sungrae and Chun, Sanghyuk},
    booktitle = {Computer Vision -- ECCV 2022},
    pages = {440--457},
    year = {2022},
    doi = {10.1007/978-3-031-20050-2_26}
}

@inproceedings{bridle1990probabilistic,
    title = {Probabilistic Interpretation of Feedforward Classification Network Outputs, with Relationships to Statistical Pattern Recognition},
    author = {Bridle, John S},
    booktitle = {Neurocomputing: Algorithms, Architectures and Applications},
    pages = {227--236},
    year = {1990},
    doi = {10.1007/978-3-642-76153-9_28}
}

@inproceedings{robey2021model,
    title = {Model-Based Domain Generalization},
    author = {Robey, Alexander and Pappas, George J. and Hassani, Hamed},
    booktitle = {Advances in Neural Information Processing Systems},
    volume = {34},
    pages = {20210--20229},
    year = {2021},
}

@inproceedings{foster2019hypothesis,
    title = {Hypothesis Set Stability and Generalization},
    author = {Foster, Dylan J and Greenberg, Spencer and Kale, Satyen and Luo, Haipeng and Mohri, Mehryar and Sridharan, Karthik},
    booktitle = {Advances in Neural Information Processing Systems},
    volume = {32},
    year = {2019},
}

@inproceedings{senderovich2022towards,
    title = {Towards Practical Control of Singular Values of Convolutional Layers},
    author = {Senderovich, Alexandra and Bulatova, Ekaterina and Obukhov, Anton and Rakhuba, Maxim},
    booktitle = {Advances in Neural Information Processing Systems},
    volume = {35},
    year = {2022},
}

@inproceedings{pan2021disentangled,
    title = {Disentangled Information Bottleneck},
    author = {Pan, Ziqi and Niu, Li and Zhang, Jianfu and Zhang, Liqing},
    booktitle = {Proceedings of the AAAI Conference on Artificial Intelligence},
    volume = {35},
    pages = {9285--9293},
    year = {2021},
    doi = {10.1609/aaai.v35i10.17120}
}

@inproceedings{brock2021high,
    title = {High-Performance Large-Scale Image Recognition Without Normalization},
    author = {Brock, Andy and De, Soham and Smith, Samuel L and Simonyan, Karen},
    booktitle = {Proceedings of the 38th International Conference on Machine Learning},
    series = {Proceedings of Machine Learning Research},
    volume = {139},
    pages = {1059--1071},
    year ={2021},
}

@inproceedings{kalatzis2020variational,
    title = {Variational Autoencoders with {R}iemannian Brownian Motion Priors},
    author = {Kalatzis, Dimitris and Eklund, David and Arvanitidis, Georgios and Hauberg, S{\o}ren},
    booktitle = {Proceedings of the 37th International Conference on Machine Learning},
    series = {Proceedings of Machine Learning Research},
    volume = {119},
    pages = {5053--5066},
    year = {2020},
}

@inproceedings{long2019generalization,
    title = {Generalization bounds for deep convolutional neural networks},
    author = {Philip M. Long and Hanie Sedghi},
    booktitle = {International Conference on Learning Representations},
    year = {2019}
}

@inproceedings{sedghi2018singular,
    title = {The Singular Values of Convolutional Layers},
    author = {Sedghi, Hanie and Gupta, Vineet and Long, Philip M},
    booktitle = {International Conference on Learning Representations},
    year = {2018}
}

@inproceedings{brock2021characterizing,
    title = {Characterizing signal propagation to close the performance gap in unnormalized resnets},
    author = {Brock, Andrew and De, Soham and Smith, Samuel L},
    booktitle = {International Conference on Learning Representations},
    year = {2021}
}

@article{hendrycks2016baseline,
  title={A baseline for detecting misclassified and out-of-distribution examples in neural networks},
  author={Hendrycks, Dan and Gimpel, Kevin},
  journal={arXiv preprint arXiv:1610.02136},
  year={2016}
}

@inproceedings{cai2020real,
  title={Real-time out-of-distribution detection in learning-enabled cyber-physical systems},
  author={Cai, Feiyang and Koutsoukos, Xenofon},
  booktitle={2020 ACM/IEEE 11th International Conference on Cyber-Physical Systems (ICCPS)},
  pages={174--183},
  year={2020},
  organization={IEEE}
}

@misc{bartlett2013theoretical,
    title = {Theoretical Statistics Lecture 14},
    author = {Bartlett, Peter},
    year = {2013},
    howpublished = "\url{https://www.stat.berkeley.edu/~bartlett/courses/2013spring-stat210b/notes/14notes.pdf}",
    note = "Accessed: 2023-08-08",
}

@inproceedings{Dosovitskiy2017,
  title = 	 {{CARLA}: {An} Open Urban Driving Simulator},
  author = 	 {Dosovitskiy, Alexey and Ros, German and Codevilla, Felipe and Lopez, Antonio and Koltun, Vladlen},
  booktitle = 	 {Proceedings of the 1st Annual Conference on Robot Learning},
  pages = 	 {1--16},
  year = 	 {2017},
  editor = 	 {Levine, Sergey and Vanhoucke, Vincent and Goldberg, Ken},
  volume = 	 {78},
  series = 	 {Proceedings of Machine Learning Research},
  month = 	 {13--15 Nov},
  publisher =    {PMLR},
  url = 	 {https://proceedings.mlr.press/v78/dosovitskiy17a.html}
}

@ARTICLE{9126102,
  author={Cass, Stephen},
  journal={IEEE Spectrum}, 
  title={Nvidia makes it easy to embed AI: The Jetson nano packs a lot of machine-learning power into DIY projects - [Hands on]}, 
  year={2020},
  volume={57},
  number={7},
  pages={14-16},
  doi={10.1109/MSPEC.2020.9126102}}

@inproceedings{vasilev2020q,
  title={q-Space novelty detection with variational autoencoders},
  author={Vasilev, Aleksei and Golkov, Vladimir and Meissner, Marc and Lipp, Ilona and Sgarlata, Eleonora and Tomassini, Valentina and Jones, Derek K and Cremers, Daniel},
  booktitle={Computational Diffusion MRI: MICCAI Workshop, Shenzhen, China, October 2019},
  pages={113--124},
  year={2020},
  organization={Springer}
}

@article{zhang2020towards,
  title={Towards out-of-distribution detection with divergence guarantee in deep generative models},
  author={Zhang, Yufeng and Liu, Wanwei and Chen, Zhenbang and Wang, Ji and Liu, Zhiming and Li, Kenli and Wei, Hongmei},
  journal={arXiv preprint arXiv:2002.03328},
  year={2020}
}

\newpage

\onecolumn

\title{Disentangled and Distilled Encoder for Out-of-Distribution Reasoning with Rademacher Guarantees\\(Supplementary Material)}
\maketitle

\appendix
\section{Details of optimality analysis}

This section presents the required assumptions and propositions for analyzing the optimality of solutions of a defined constrained optimization problem.

\subsection{Feasibility assumptions}\label{A:FAss}

\begin{assumption}[Feasibility condition for problem ~\ref{P:p2}]\label{as:feasprob4}
    For encoder model $\mathcal{E}_s$, there is a parameter $\theta_s \in \Theta_s$ that satisfies disentanglement constraints:

    \begin{equation}
    \begin{aligned}
       & \frac{1}{m_A} \sum_{i=1}^{m_A}  \mathcal{L}^{\diamond}_{A,f} (x_i,x'_i) \leq \gamma_{A,f} - \xi \\
       &\frac{1}{m_I} \sum_{i=1}^{m_I}  \mathcal{L}^{\diamond}_{I,f} (x_i,x'_i) \leq \gamma_{I,f} - \xi
    \end{aligned}  
    \end{equation}
    where  $\mathcal{L}^{\diamond}_{A,f},\mathcal{L}^{\diamond}_{I,f}$ are Lipschitz and bounded losses that are defined in Section ~\ref{sec: TR} and   $\xi >0$.
    \label{as:feasp1}
\end{assumption}

\begin{assumption}[Feasibility condition for Problem 3 of Table ~\ref{TB:OPTPD}]\label{as:feasprob7}
    For encoder model $\mathcal{E}_s$, there is a parameter $\theta'_s \in \Theta_s$ that satisfies disentanglement constraints:

    \begin{equation}
    \begin{aligned}
       & E_{(x,x') \sim \mathfrak{D}(x)}\mathcal{L}^{\diamond}_{A,f} (x_i,x'_i) \leq \gamma_{A,f} - \kappa_A \epsilon - \xi\\
        &E_{(x,x') \sim \mathfrak{D}(x)} \mathcal{L}^{\diamond}_{I,f} (x_i,x'_i) \leq \gamma_{I,f} - \kappa_I \epsilon - \xi
    \end{aligned}  
    \end{equation}
      Where $\mathcal{L}^{\diamond}_{A,f}$ and $\mathcal{L}^{\diamond}_{I,f}$ are defined in Section ~\ref{sec: TR} and are   Lipschitz and bounded losses.   $\xi >0$ and $\epsilon$ is the maximum of $\epsilon_{\tau} \; and \;  \epsilon_s  \;$  that are defined in assumptions  ~\ref{as:Compteach} and ~\ref{as:lowcom}, respectively.
    \label{as:feasp2}
\end{assumption}
\subsection{Required assumptions for proposition ~\ref{prp:gbtt}}\label{sbc:RCas}

The following assumptions are defined for distillation loss to obtain $\kappa_d\omega$-stability in proposition ~\ref{prp:gbtt}. However, the same assumptions for adaptation and isolation losses can be defined to obtain  $\kappa_A\omega$-stability and $\kappa_I\omega$-stability, respectively.
\begin{assumption}[$\omega$-sensitivity of teacher model] \label{as:sens}
 The  teacher model $\mathcal{E}_{\tau}:\mathcal{X}_{T} \longrightarrow Z_{\tau} $ is $\omega$-sensitive for training samples $\mathcal{X}_{T}$ and $\mathcal{X}_{T'}$ that only differ in one sample.

\begin{equation}
    \begin{aligned}
        &\forall  \mathcal{X}_{T} ,\exists  \mathcal{X}'_{T}:
    \forall x \in  \mathcal{X}_{T},\forall x'  \in \mathcal{X}'_{T}: ||\mathcal{E}_{\tau}(\theta_{\tau},x)-\mathcal{E}_{\tau}(\theta_{\tau},x')||_{\infty} \leq \omega
    \end{aligned}
\end{equation}
\label{as:omegtech}
\end{assumption}
\begin{assumption}[Stability of student hypothesis space]\label{as:stbstd}
    The student hypothesis must have the following characteristics to ensure that the obtained loss by the student hypothesis does not drastically change by substituting one sample with another during training. 
    Student hypothesis space $\mathcal{H}_s$ has   CV-stability $\Upsilon$, average  CV-stability  $\overline{\Upsilon}$, and maximum diameter  $\Upsilon_{Max}$ ~\citep{foster2019hypothesis}:  
     \begin{itemize}
         \item CV-stability:
          \begin{equation}
         \begin{aligned}
             &sup_{x \in X_{T}} E_{x' \in \mathcal{X} \setminus \mathcal{X}_{T},x \in \mathcal{X}_{T}}[sup_{\theta_s ,\theta'_s \in \Theta_s}  [JS \circ SF(\mathcal{E}_{\tau}(\theta_{\tau},x),\mathcal{E}_s(\theta_s,x))-
             & \textsc{JS} \circ \textsc{SF}(\mathcal{E}_{\tau}(\theta_{\tau},x),\mathcal{E}_s(\theta'_s,x))]]  \leq \Upsilon
         \end{aligned}
     \end{equation}
    Where $\mathcal{X}'_{T}$ is a training set with sample $x$ from $\mathcal{X}_{T}$ is replaced by $x'$. $\theta_s \; and \; \theta'_s$ are parameters of the encoder that are learned by using training samples $\mathcal{X}_{T}$ and $\mathcal{X}'_{T}$, respectively.
     \item Average CV-stability:
        \begin{equation}
         \begin{aligned}
             &E_{X_{T} \subset \mathcal{X} } E_{x' \in \mathcal{X} \setminus \mathcal{X}_{T},x \in \mathcal{X}_{T}}[\textsc{JS} \circ \textsc{SF}(\mathcal{E}_{\tau}(\theta_{\tau},x),\mathcal{E}_s(\theta_s,x))-  \textsc{JS} \circ \textsc{SF}(\mathcal{E}_{\tau}(\theta_{\tau},x),\mathcal{E}_s(\theta'_s,x))] \leq \overline{\Upsilon}
         \end{aligned}
     \end{equation}
     \item Maximum diameter:
     \begin{equation}
         \begin{aligned}
           &sup_{x \in X_{T}}max_{x}[sup_{\theta_s ,\theta'_s \in \Theta} [\textsc{JS} \circ \textsc{SF}(\mathcal{E}_{\tau}(\theta_{\tau},x),\mathcal{E}_s(\theta_s,x))-  \textsc{JS} \circ \textsc{SF}(\mathcal{E}_{\tau}(\theta_{\tau},x),\mathcal{E}_s(\theta'_s,x))]] \leq \Upsilon_{Max}
         \end{aligned}
     \end{equation}

     \end{itemize}   
    \label{as:stabst}

\end{assumption}
Assumptions ~\ref{as:omegtech} and  ~\ref{as:stabst} indicate that a slight change in training data for teacher and student encoders does not significantly change their outputs. These assumptions generally hold when a model is adequately trained with tuned hyperparameters.

\begin{assumption}\label{as:Lippar}
    The loss function  $\mathcal{L}^{\diamond}_{D}(x)$ is Lipschitz parameterized:
    \begin{equation}
         \begin{aligned}
            \forall \theta_s \in \Theta_s:
            ||\frac{\delta \textsc{JS} \circ \textsc{SF}(\mathcal{E}_{\tau}(\theta_{\tau},x),\mathcal{E}_s(\theta_s,x))}{\delta \theta_s}||_p \leq \kappa_{\theta}
         \end{aligned}
     \end{equation}
     Here the $||.||_p$ is p-norm. 
     \label{as:parlips}
\end{assumption}
The encoder model is Lipschitz parameterized by controlling the Lipschitz coefficient of the student encoder using the approach introduced in section ~\ref{scb:RCcont}.

\begin{assumption}\label{as:cls}
The  student hypothesis space $\mathcal{H}_s$ includes only student models that are $\gamma_c$-close to teacher model: 

    \begin{equation}
        \begin{aligned}   
     &\mathcal{H}_s=\{\mathcal{E}_s| \; ||\mathcal{E}_{\tau}(\theta_{\tau},x)-\mathcal{E}_s(\theta_s,x)||_{\infty} \leq \gamma_c\}
        \end{aligned}
    \end{equation}
    \label{as:close}
\end{assumption}

\begin{proposition}[$\kappa_D \omega-$ stability of student hypothesis space (from section 5.4 of ~\citep{foster2019hypothesis})] \label{lm:stb}
Consider   teacher models $\mathcal{E}_{\tau}=\mathcal{E}_{\tau}(\theta_{\tau},x)$  and  $\mathcal{E}'_{\tau}=\mathcal{E}_{\tau}(\theta'_{\tau},x')$ that are trained  with $x \in \mathcal{X}_{T}=\{x_1,..,x_j, ..., x_m\}$  and $x' \in \mathcal{X}'_{T}=\{x_1,..,x'_j, ..., x_m\}$, respectively. $\mathcal{X}_{T}$ and $ \mathcal{X}'_{T}$ only differ in one sample. Consider $\mathcal{E}_{\tau}$ and $\mathcal{E}'_{\tau}$ are not  in $\mathcal{H}_s$, but $||\mathcal{E}_{\tau}-\mathcal{E}'_{\tau}||_{\infty}  \in  \mathcal{H}_s$. When  $\mathcal{E}_s \in \mathcal{H}_s$,  as  student models are $\gamma_c$-close to their respective teacher models  ($||\mathcal{E}_s(\theta_s,x) - \mathcal{E}_{\tau}(\theta_{\tau},x)||_{\infty}=||\mathcal{E}'_s(\theta_s',x') -\mathcal{E}'_{\tau}((\theta'_{\tau},x'))||_{\infty} \leq \gamma_c$), then $\mathcal{E}'_s=\mathcal{E}_s+\mathcal{E}_{\tau} - \mathcal{E}'_{\tau} \in \mathcal{H}'_s$,  where $ \mathcal{H}'_s$
is the hypothesis space  obtained by training with $\mathcal{X}'_{T}$.
\begin{equation}
\begin{aligned}
&|[\textsc{JS} \circ \textsc{SF}(\mathcal{E}_{\tau}(\theta_{\tau},x),\mathcal{E}_s(\theta_s,x))- 
              \textsc{JS} \circ \textsc{SF}(\mathcal{E}_{\tau}(\theta_{\tau},x),\mathcal{E}_s(\theta'_s,x))]|  \\
              &\leq \kappa_D *|\mathcal{E}_s(\theta_s,x)-\mathcal{E}_s(\theta'_s,x')| =
\kappa_D *|\mathcal{E}_\tau(\theta_{\tau},x)-\mathcal{E}_{\tau}(\theta'_{\tau},x')| \leq \kappa_D \omega 
\end{aligned}
\end{equation}
\label{prp:stb}
\end{proposition}
By considering assumptions in  Appendix ~\ref{sbc:RCas}, we can show $\kappa_A \omega-$ stability and $\kappa_I \omega-$ stability of student hypothesis space for adaptation and isolation losses, respectively.

\subsection{Bound over expectation of adaptation and isolation losses} \label{app:RC}
 $\kappa_A\omega-$stability and  $\kappa_I\omega-$stability of the student hypothesis can be established based on Assumption  ~\ref{as:cls} and the Lipschitzness of    
  $\mathcal{L}^{\diamond}_A$ and  $\mathcal{L}^{\diamond}_I$  by redefining proposition ~\ref{prp:stb} for these losses. 
Then by applying proposition ~\ref{prp:gbtt}, Talagrand's lemma and Dudley theorem, we obtain:
\begin{equation}
    \begin{aligned}
     &\zeta_{A}=2*\chi*\kappa_A*\Delta_{OP}*(1+\nu+\frac{\Delta_{OP}}{L})^L \sqrt{\frac{8.7*d}{m_A}}+(B_A+2\kappa_A \omega m_A)*\sqrt{\frac{1}{2m_A}ln \frac{1}{\delta}} \\
     &\zeta_{I}=2*\chi*\kappa_I*\Delta_{OP}*(1+\nu+\frac{\Delta_{OP}}{L})^L \sqrt{\frac{8.7*d}{m_I}}+(B_I+2\kappa_I \omega m_I)*\sqrt{\frac{1}{2m_I}ln \frac{1}{\delta}}
    \end{aligned}
    \label{eq:allzetas}
\end{equation}

Here, $m_I$ and $m_I$ are sizes of a subset of training space that
is used for adaptation loss and isolation loss, respectively.
$B_A$ and $B_I$ are bounds over a range of adaptation and isolation losses, respectively. Also, $\kappa_A$ and $\kappa_I$ are the Lipschitz coefficients for adaptation and isolation losses, respectively.


\section{Implementation details}

\subsection{Data generation and partitions}\label{App:Datgen}
We use the same data and partitions presented in the WDLVAE for a fair comparison between the teacher and student models. The selected generative factors are rain (\textsc{R}) and background (\textsc{BK}) ($\mathcal{F}=\{R, BK\}$), and we obtained data partitions by combining different values for these factors. For rain factor we change rain intensity from   $[0, 0]$ (NR), $[0.002, 0.003]$ (LR), $[0.005, 0.006]$ (MR) and  $[0.008, 0.009]$ (HR). For gathering different values for the background generative factor, we drive a car in the CARLA simulator in cities three ($SC3$), four ($SC4$), and five ($SC5$). Cities three, four, and five are images of rural roads, highways, and urban roads. We obtain data partitions by combining different values for these factors. Table ~\ref{TB:parts} shows data partitions, the observed values for rain and background factors in each partition, and the number of samples in those partitions in the training, validation, and test sets. Note that training, validation, and test sets are mutually exclusive.
 To avoid bias in the AUROC of rain and background reasoners, we select an equal number of ID and OOD samples in the test sets ~\citep{hendrycks2016baseline}.


\begin{table}[]
\caption{Data partitions and number of samples from partitions in training, validation and test datasets}
\resizebox{\textwidth}{!}{%
\begin{tabular}{|c|c|c|c|c|cc|}
\hline
\multirow{2}{*}{\textbf{Partition}} & \multirow{2}{*}{\textbf{Background}} & \multirow{2}{*}{\textbf{Rain}} & \multirow{2}{*}{\textbf{Train}} & \multirow{2}{*}{\textbf{Validation}} & \multicolumn{2}{c|}{\textbf{Test}} \\ \cline{6-7} 
 &  &  &  &  & \multicolumn{1}{c|}{\textbf{Rain reasoner}} & \textbf{Background reasoner} \\ \hline
P1 & SC3(City 3) & LR($[0.002, 0.003]$) & 750 & 150 & \multicolumn{1}{c|}{324} & 81 \\ \hline
P2 & SC3 (City 3) & MR($[0.005, 0.006]$) & 750 & 150 & \multicolumn{1}{c|}{324} & 81 \\ \hline
P3 & SC4 (City 4) & LR($[0.002, 0.003]$) & 750 & 150 & \multicolumn{1}{c|}{324} & 81 \\ \hline
P4 & SC4 (City 4) & MR($[0.005, 0.006]$) & 750 & 150 & \multicolumn{1}{c|}{324} & 81 \\ \hline
P5 & SC3 (City 3) & HR ($[0.008, 0.009]$) & 0 & 0 & \multicolumn{1}{c|}{162} & 81 \\ \hline
P6 & SC3 (City 3) & NR($[0,0]$) & 0 & 0 & \multicolumn{1}{c|}{162} & 81 \\ \hline
P7 & SC4 (City 4) & HR ($[0.008, 0.009]$) & 0 & 0 & \multicolumn{1}{c|}{162} & 81 \\ \hline
P8 & SC4 (City 4) & NR($[0,0]$) & 0 & 0 & \multicolumn{1}{c|}{162} & 81 \\ \hline
P9 & SC5 (City 5) & LR($[0.002, 0.003]$) & 0 & 0 & \multicolumn{1}{c|}{162} & 162 \\ \hline
P10 & SC5 (City 5) & MR($[0.005, 0.006]$) & 0 & 0 & \multicolumn{1}{c|}{162} & 162 \\ \hline
P11 & SC5 (City 5) & HR ($[0.008, 0.009]$) & 0 & 0 & \multicolumn{1}{c|}{162} & 162 \\ \hline
P12 & \multicolumn{1}{l|}{SC5 (City 5)} & NR($[0,0]$) & 0 & 0 & \multicolumn{1}{c|}{162} & 162 \\ \hline
\end{tabular}%
}

\label{TB:parts}
\end{table}

\subsection{Teacher Architecture}\label{scb:Techach}
We use   \textsc{WDLVAE}   as a teacher model with five convolution layers $32/64/128/256/512$ with kernel size  $3$, stride  $2$, and padding  $1$. Each layer is followed by batch normalization and Leaky ReLU activation function. The latent space size is $N=30$. The decoder is a mirror architecture of the encoder.
\subsection{Controlling the Rademacher Complexity of the Student Encoder in Practice:} \label{scb:RCcont}

The following approach is used to control the RC of the model in practice.
Based on Definition~\ref{df:NNLips}, during the training of the encoder, the singular values of each layer should be bounded to control the Lipschitz coefficient of the layer. It is time-consuming to find the singular values of a convolution operation by applying SVD  ~\citep{hoecker1996svd} on its corresponding circulant matrix. A more efficient method is to decompose the circulant matrix of a convolutional filter into three lower-ranked matrices ~\citep{senderovich2022towards} using tensor train (TT) decomposition ~\citep{zniyed2022structured}. The first and last matrices are orthogonal, while the middle matrix with rank {\fontsize{8}{8}\selectfont $d_D$} has the same singular values as the original matrix. Since the singular vectors of the circulant matrix are Fourier basis vectors ~\citep{sedghi2018singular}, the Fourier coefficient of the convolution filter is calculated. Then, SVD is applied to the middle lower-ranked matrix that is obtained from TT decomposition. We use clipping to bound the values of the singular values of each layer. Clipping involves replacing the singular values of linear operations corresponding to a convolution layer that exceeds a predefined threshold with that threshold: ( $(\forall  \textsc{SN(OP(R))}: \textsc{SN(OP(R))} \ge \vartheta \longrightarrow \textsc{SN(OP(R))} = \vartheta$). Here $\textsc{SN}$ is the function that extracts singular values, and $\vartheta $ is a predefined threshold.

Table~\ref{tb:assCOP} shows the assigned values for defining the disentangled distilled student encoder and input of  Algorithm ~\ref{alg:algs}. 
Also, we select gain coefficient $\Gamma=1.7$  for all layers to normalize their variance to one. We also select $d_D=400$ as the decreasing rank for the middle matrix in TT decomposition.
\begin{table}[]
\caption{Values assigned to different variables for defining training as constraint optimization and inputs of Algorithm \ref{alg:algs}.}
\label{tb:assCOP}
\centering
\resizebox{0.4\textwidth}{!}{%
\begin{tabular}{|c|c|c|c|}
\hline
\textbf{Variable} & \textbf{Value} & \textbf{Variable} & \textbf{Value} \\ \hline
$m$                               & 3000   & $\lambda^0_{A,R}$                       & 2   \\ \hline
$m_A$                             & 1500   &  $\lambda^0_{I,R}$                       & 2 \\ \hline
$m_I$                             & 1500   &  $\lambda^0_{A,BK}$                      & 10 \\ \hline
$\gamma_{A,R}$                    & 0.1   &  $\lambda^0_{I,BK}$                      & 10   \\ \hline
$\gamma_{A,BK}$ & 0.0001   &  $\eta_D$                                & 0.00001   \\ \hline
$\gamma_{I,R}$                    & 0.1   & $\eta_{A,R}$                            & 0.05     \\ \hline
$\gamma_{I,BK}$                   & 0.0001   &  $\eta_{I,R}$                            & 0.05     \\ \hline
$\mathcal{Z}^s_{R}$ & \{3\}  &  $\eta_{A,BK}$                           & 0.5   \\ \hline
$\mathcal{Z}^s_{BK}$ & \{6\}   &  $\eta_{I,BK}$                           & 0.5  \\ \hline
 $|\mathcal{Z}^{s}_{R}|$                   & 1  & \multicolumn{2}{l}{\multirow{2}{*}{}} \\ \cline{1-2}
 $|\mathcal{Z}^s_{BK}|$                   & 1  & \multicolumn{2}{l}{} \\ \cline{1-2}

\end{tabular}%
}

\end{table}

\subsection{Time measurements on Jetson Nano}\label{scb:TM}
To measure the timing and memory usage of our student and teacher models, we used a Jetson Nano~\citep{9126102}, a low-power compute unit designed for inferencing neural networks that have been deployed in many robotic applications. Table~\ref{tb:jetson} shows the hardware and software configuration used in our experiments. Network time protocol (NTP) was disabled to prevent OS clock adjustments while measuring timing data.

To measure execution time, we looped through a sequence of $1000$ images stored on the Jetson's SD card. Each image was loaded by the Python interpreter as a Pillow Image object, and the \textit{Resize} and \textit{ToTensor} transforms were applied before model inference. The inference time was measured using the OS clock, which is accurate to $\pm1 \mu$s. To measure memory usage, we considered the cumulative size of all the tensors in the model stored with 32-bit floating point precision.
\begin{table}[hbt!]
\caption{Hardware and software setup for timing and memory consumption experiments.}
\label{tb:jetson}
\centering
\begin{tabular}{|c|c|}
\multicolumn{2}{c}{\textbf{Hardware}} \\ \hline
\textit{CPU Type}        & ARM Cortex-A57      \\ \hline
\textit{CPU Core Count}  & 4                   \\ \hline
\textit{CPU Clock Speed} & 1.479 GHz           \\ \hline
\textit{Memory}          & 2GB DDR4            \\ \hline
\multicolumn{2}{c}{\textbf{Software}} \\ \hline
\textit{OS}              & L4T 32.1            \\ \hline
\textit{PyTorch Version} & 1.8                 \\ \hline
\end{tabular}

\end{table}


\section{Additional results}\label{scb:appres}
In this section, additional results of experiments are presented. We use a student encoder with a compression rate of $0.5$, which provides good OOD performance for these experiments.


\subsection{Satisfaction of disentanglement  losses}

\begin{figure}[hbt!]
    \centering
    \includegraphics[width=\textwidth]{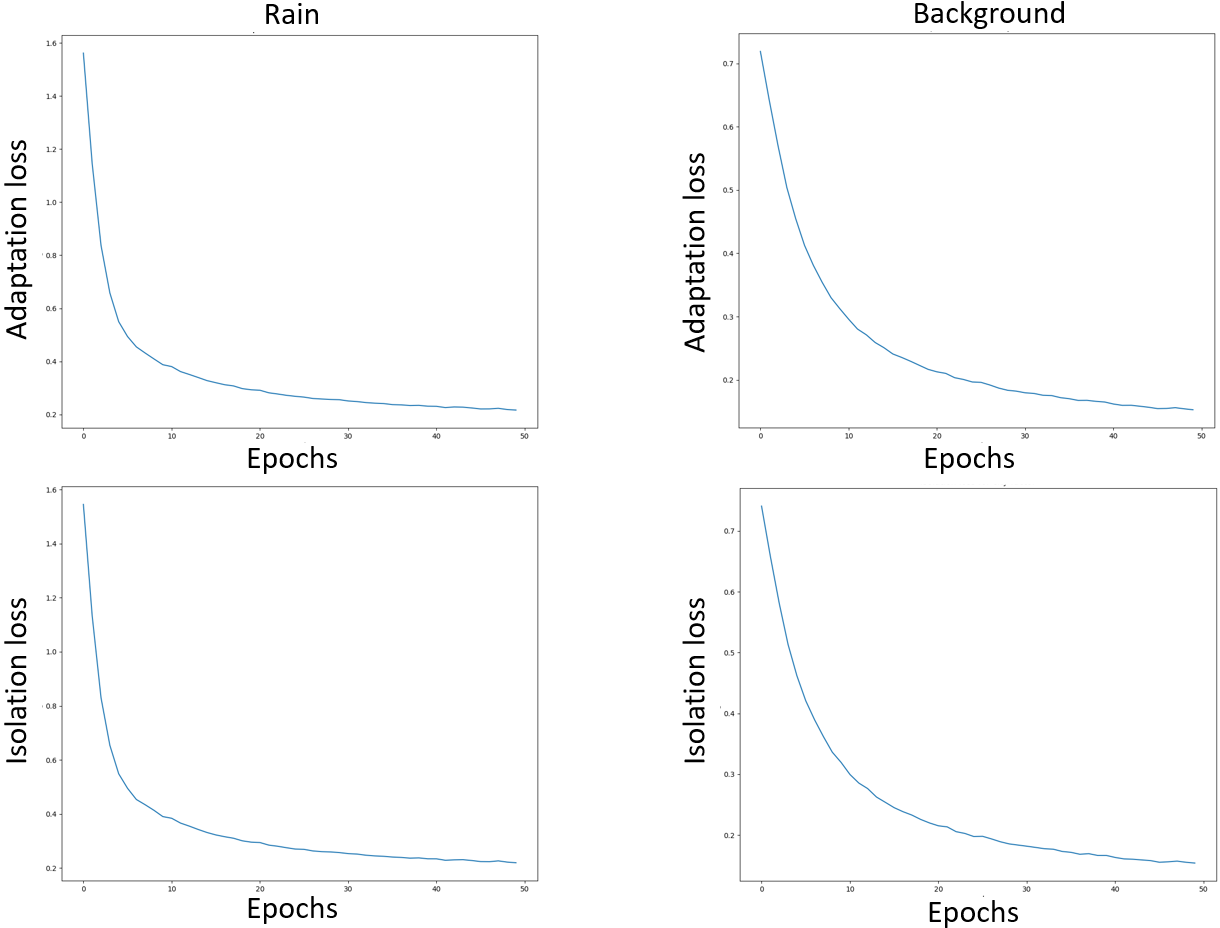}
    \caption{Satisfaction of adaptation and isolation losses to predefined margins for compression rate $0.5$}
    \label{fig:losses5}
\end{figure}

Figure ~\ref{fig:losses5} shows the satisfaction of adaptation and isolation losses for rain and background factors. Increasing the epochs decreases the value of losses, and they converge to margin variables.

\subsection{Calculating Rademacher complexity and Rademacher plots}

\begin{figure}
    \centering
\includegraphics[width=\textwidth]{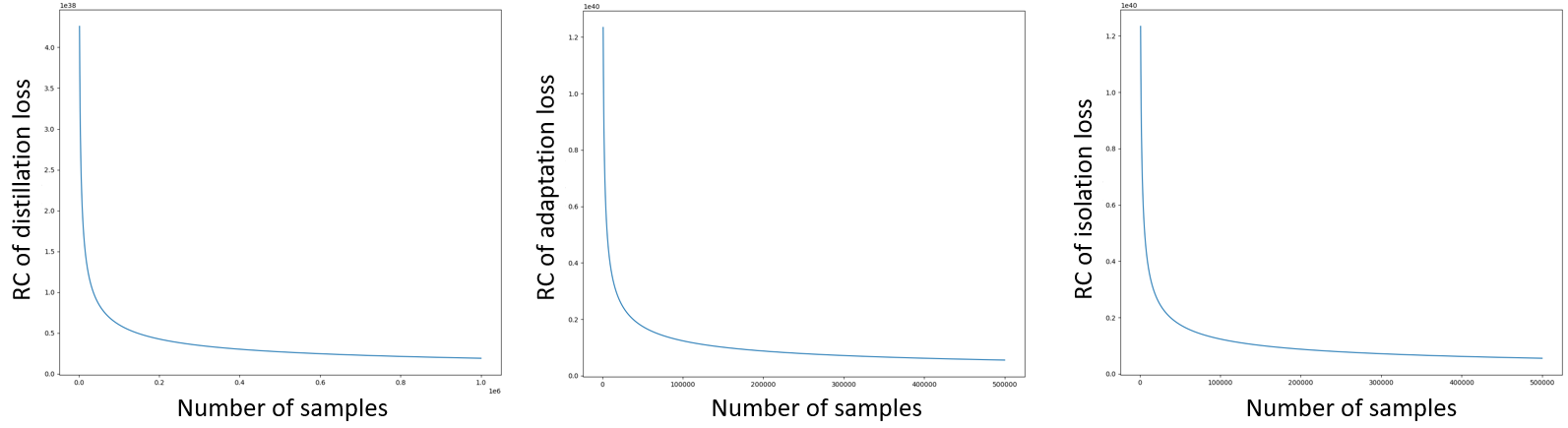}

    \caption{Rademacher complexity of distillation, adaptation and isolation losses for rate 0.5 compression.}
    \label{fig:RC5}
\end{figure}

Table~\ref{tb:RDvals} values are used to calculate Rademacher complexities.
Figure ~\ref{fig:RC5}  shows the Rademacher complexity of distillation, adaptation, and isolation losses for a compression rate  $0.5$. As shown in the figures, the RC is decreasing function with respect to sample size for all loss functions and is well defined.
\begin{table}[]
\caption{Values of variables required to calculate Rademacher complexity.}
\label{tb:RDvals}
\centering
\resizebox{0.25\textwidth}{!}{%
\begin{tabular}{|c|c|}
\hline
\textbf{Variable} & \textbf{Value} \\ \hline
$\kappa_D$        & 3              \\ \hline
$\kappa_A$        & 61.5           \\ \hline
$\kappa_I$        & 206.4          \\ \hline
$B_D$             & 1              \\ \hline
$B_A$             & 54.72          \\ \hline
$B_I$             & 216.8          \\ \hline
$L$             & 7          \\ \hline
$\omega$             & 0.001          \\ \hline
$\delta$             & 0.1          \\ \hline
$\chi$ &  2519 \\ \hline
$d$& 1144752  \\ \hline
$1+\nu$ &4899 \\ \hline 
 $\Delta_{OP}$ & 15920   \\ \hline 
\end{tabular}%
}

\end{table}
      

\end{document}